\documentclass[11pt]{article}

\usepackage[utf8]{inputenc}
\usepackage{hyperref}
\usepackage{graphicx}
\usepackage{subcaption}
\usepackage{amsmath}
\usepackage{multirow}
\usepackage{rotating}
\usepackage{xspace}
\usepackage{amssymb}
\usepackage{pifont}
\usepackage{authblk}  
\usepackage[margin=0.85in]{geometry}

\newcommand{\datasetname}{NFI\_FARED\xspace}
\newcommand{\modelname}{Hi-OSCAR\xspace}

\title{\modelname: Hierarchical Open-set Classifier for Human Activity Recognition}

\author[1]{Conor McCarthy}
\author[2]{Loes Quirijnen}
\author[2,3]{Jan Peter van Zandwijk}
\author[2,1]{Zeno Geradts}
\author[1]{Marcel Worring}

\affil[1]{University of Amsterdam, Amsterdam, The Netherlands}
\affil[2]{Netherlands Forensic Institute (NFI), The Hague, The Netherlands}
\affil[3]{Amsterdam University of Applied Sciences, Amsterdam, The Netherlands}

\date{}

\begin{document}

\maketitle

\begin{abstract}
Within Human Activity Recognition (HAR), there is an insurmountable gap between the range of activities performed in life and those that can be captured in an annotated sensor dataset used in training. Failure to properly handle unseen activities seriously undermines any HAR classifier's reliability. Additionally within HAR, not all classes are equally dissimilar, some significantly overlap or encompass other sub-activities. Based on these observations, we arrange activity classes into a structured hierarchy. From there, we propose \modelname: a \textbf{Hi}erarchical \textbf{O}pen-set \textbf{C}lassifier for \textbf{A}ctivity \textbf{R}ecognition, that can identify known activities at state-of-the-art accuracy while simultaneously rejecting unknown activities. This not only enables open-set classification, but also allows for unknown classes to be localized to the nearest internal node, providing insight beyond a binary "known/unknown" classification. To facilitate this and future open-set HAR research, we collected a new dataset: \datasetname. \datasetname contains data from multiple subjects performing nineteen activities from a range of contexts, including daily living, commuting, and rapid movements, which is fully public and available for download\footnote{\url{https://osf.io/sczrh/}}. 
\end{abstract}

\section*{Introduction} \label{section:introduction}
Human Activity Recognition (HAR) is the process of classifying activities using data collected from one or more sensors placed either on the body or in the environment where the activity is taking place. With the continuous improvement of sensor technology, body-worn HAR has received a lot of attention because of its potential scalability, high privacy, and relatively low cost when compared to using cameras and other environment-based sensors. Body-worn HAR has found applications in sports \cite{jung_lax-score_2021, jia_swingnet_2021}, education \cite{gao_individual_2022, gao_n-gage_2020}, medicine \cite{wang_leveraging_2021, chen_apneadetector_2021, bartolome_glucomine_2021}, smart homes \cite{arrotta_dexar_2022}, and forensics \cite{jennings_interpreting_2023, van_zandwijk_iphone_2019, van_zandwijk_phone_2021}. Apart from the lower cost and wider selection of sensors being  available now, the continued advancement of AI-based techniques has improved classification performance and added capabilities.
\par
AI techniques for HAR rely on large annotated datasets. A limiting factor for practical applications is the coverage gap between datasets for training annotated with a limited number of activities and real-world activity variations. In any uncontrolled setting, the number of encountered activities will greatly outnumber the set of classes in the data used for training. Datasets are needed that are taking in real-world conditions and are annotated with a large variety of activities. But even when AI tools are trained with a large number of activities, there will always be activities that have not yet been encountered. Failure to appropriately flag and process such unknown activities can render the output of activity classification models worthless. Consequently, an area of HAR seeing increased attention is open-set classification. Open-set classification models are able to label samples as Out-of-Distribution (OOD) if the class is from outside the training set. In contrast, closed-set classification restricts the model to only recognise In-Distribution (ID) classes present in the training set \cite{chen_open-set_2024, jain_multi-class_2014}. A closed-set approach forces the model to erroneously classify OOD input as one of the classes from the training set, sometimes with high reported certainty. Given the wide array of activities that can reasonably be encountered in practical settings, open set classification is a necessary component of HAR. 
\par
\begin{figure}[ht]
    \centering
    \includegraphics[width=1.0\textwidth]{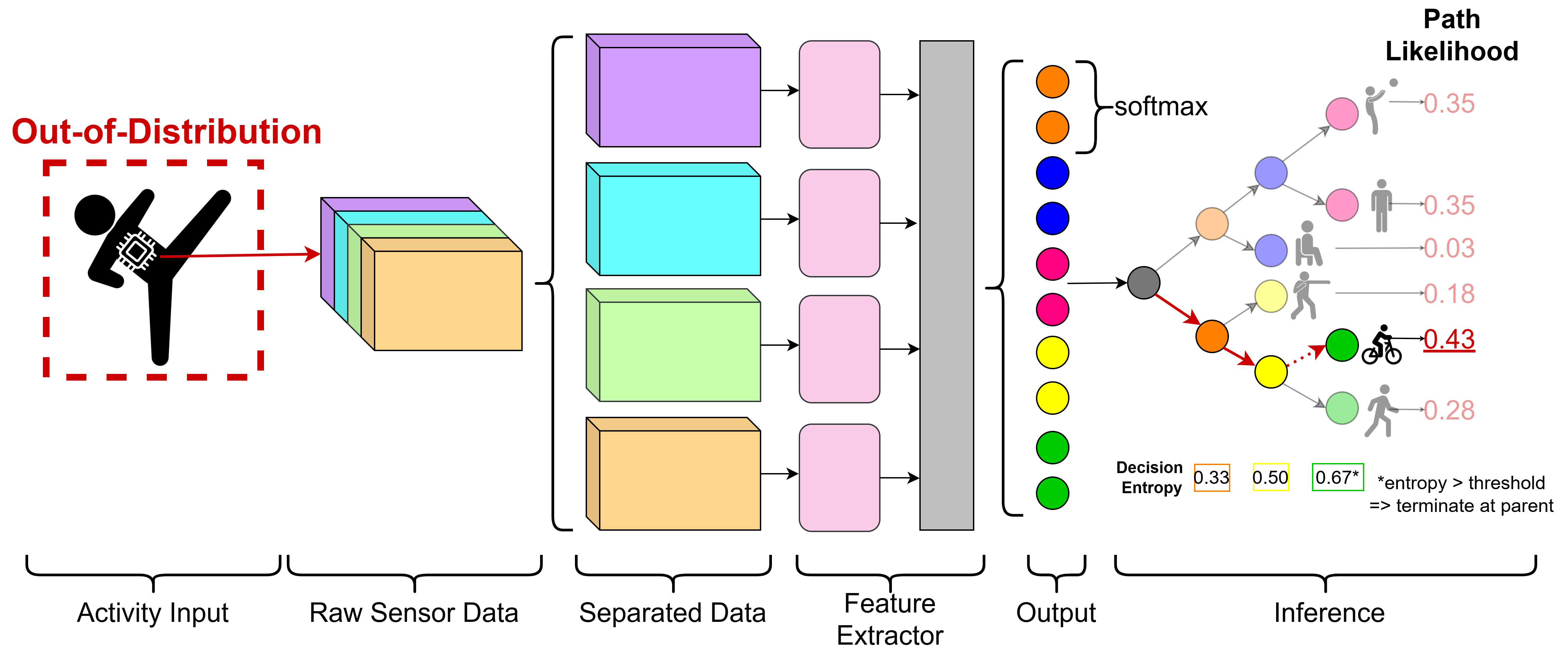}
    \caption{Model inference for an Out-of-Distribution activity using \modelname. Raw sensor data are separated sensor-wise and fed through a feature extractor, giving an output vector of length $|H|$ (where $|H|$ > number of classes). Elements of the output vector correspond with nodes in the hierarchy. The Path Likelihood of a class is computed by multiplying the softmax scores of each node along the path from root to node, and highest path likelihood determines the predicted path. If the decision entropy of any node along the chosen path exceeds the selected threshold, prediction is terminated. As an example here, the OOD class \textit{kicking} would classified as \textit{cycling} in the closed-set system, but due to high entropy it is classed as OOD. Prediction is also terminated at the penultimate node due to high decision entropy, indicating although it is OOD, it is similar to cycling and running.}
    \label{fig:main_overview}
\end{figure}

While typical approaches keep activities organised in a flat structure without leveraging inter-class relations, structuring classes as a hierarchy can be a useful method for improving open-set performance \cite{linderman_fine-grain_2022}. Arranging the classes into a hierarchical tree structure groups similar classes close to each other, with narrowly-defined classes below their  more generally-defined supersets. A hierarchy better represents HAR data, where not all activities are equally dissimilar, and some can often be subsets of the other. For example, \textit{walking} and \textit{walking upstairs} are more closely related than \textit{walking} and \textit{punching}. The hierarchy embeds additional information into predictions. Our model, \modelname, then makes predictions by traversing the tree through multiple internal classification decisions (Figure \ref{fig:main_overview}). When encountering and rejecting an OOD sample, the traversed path can still be inspected, and provide insight into which known classes are similar to this OOD activity. \modelname's open-set classification localises an OOD sample to the nearest internal node without changing the underlying hierarchy. For example, our model might localise the unknown class \textit{sitting} to an internal node close to the known class \textit{standing}. In the case of the unknown class being completely different to anything in the training set, nodes at or near the Root of the hierarchy would be returned. Such information is also insightful when interpreting the unknown activity. In our method, we  apply an activity-level hierarchy to mirror the structure of real-life HAR classes.
\par
It is important to create a hierarchy that is both semantically meaningful and aligns with the data signatures. Previous work in HAR considered a hierarchy of simple gestures or sub-activities which when performed in a sequence form a complete activity class \cite{tayyub_learning_2018, wang_human_2020}. This type of hierarchy falls short of then linking together the higher level classes in a meaningful way. This has been done in other domains which borrow existing hierarchies based on domain knowledge, such as the ImageNet dataset adopting the Wordnet hierarchy \cite{kilgarriff_review_2000}. No such hierarchy exists for HAR, and constructing one requires domain expertise without guarantee of consistency between experts due to a lack of objective rankings across classes e.g. is \textit{running} more closely related to \textit{walking} or \textit{kicking}? To ensure a consistent and meaningful class structure, we propose deriving a hierarchy based on statistical features of the classes and self-supervised clustering, .  
\par
Organized in a hierarchy or not, feature extraction is a central factor for classification performance. In the broader field of AI, Large Language Models (LLMs) have become foundational for most tasks, including those outside of the target domain. Within HAR, attempts to apply pretrained LLMs for zero-shot prediction still underperform smaller dedicated models \cite{ji_hargpt_2024}, and work on creating a dedicated LLM or transformer-based approach for HAR is still ongoing \cite{okita_towards_2023}. Convolutional networks remain state of the art in HAR, and like LLMs can be improved through transfer learning. We adapt the ResNet architecture from Yuan et al. \cite{yuan_self-supervised_2022}, which is based on a single-device accelerometer, to a multi-device, multi-sensor set up and fine tune the resulting hierarchical architecture on our data. 
\par
To produce robust classification models, HAR datasets should satisfy two conditions : a high diversity of classes, and a high number of subjects. A high diversity of classes allows for in depth OOD analysis and comparison. A high number of subjects is required so that some subjects can be used as the test set, which is essential for generalising the model to actual applications. In this paper we introduce a new HAR dataset: \datasetname, containing data of multiple subjects performing nineteen activities from a broad spectrum of contexts, including daily living, commuting, and rapid movements. 
\par
Contributions of this study include:
\begin{enumerate}
    \item An open-set activity classifier with both high in-distribution and out-of-distribution accuracy, reflecting the practical requirements of HAR.\footnote{Project GitHub: https://github.com/Con-or-McCarthy/Hi\_OSCAR}
    \item A hierarchical model architecture which both improves performance, and provides insight to out-of-distribution samples.
    \item \datasetname, a purpose-built HAR dataset containing body-worn sensor data for a diverse range of activities from many participants. 
\end{enumerate}
\par
The paper is laid out as follows: in Section \ref{section:relatedworks}, we provide a brief overview of existing HAR techniques and OOD detection methods. Section \ref{section:method} describes the proposed \modelname method, including the model architecture training paradigm. Our new dataset \datasetname is presented in Section \ref{section:our_dataset} and compared with a selection of already publicly available HAR datasets. Experimental setup and results are detailed in Section \ref{section:experiments}, with a discussion in Section \ref{section:discussion}. A brief summary, as well as potential future work and concluding remarks are provided in Section \ref{section:conclusion}.

\section{Related Work} \label{section:relatedworks}

\subsection{Human Activity Recognition}

Early research in HAR relied on handcrafted feature extraction usually based upon either the time domain \cite{berchtold_actiserv_2010, attal_physical_2015, kose_online_nodate}, the frequency domain \cite{alemayoh_new_2021, zhu_feature_2017, bayat_study_2014}, or both \cite{asim_context-aware_2020, shoaib_complex_2016, shoaib_fusion_2014}. Handcrafted features are relatively straightforward and cheap, and can be made even more powerful when used with machine learning algorithms such as support vector machines \cite{tran_human_2016}, decision trees \cite{nurwulan_human_2021}, k-nearest neighbours \cite{mohsen_human_2022}, random forests \cite{balli_human_2019}, or multilayer perceptrons \cite{randell_context_2000}. Handcrafted features, however, are not necessarily effective, and approaches can be both expert- and task-specific and therefore do not generalize. 
\par
Deep Learning (DL) based methods for HAR have advanced the field massively in recent years. Convolutional Neural Networks (CNNs) process high dimensional data efficiently and can extract features directly from the time series. Combining CNNs with Recurrent Neural Networks (RNNs) to leverage the temporal aspects of the data has been an effective approach for classification. Examples are CNN layers combined with LSTMs, BiLSTMs, GRUs, and BiGRUs \cite{ordonez_deep_2016, mekruksavanich_deep_2021} which are effective for classification, as well as other variations incorporating residual modules \cite{helmi_human_2023} and unique combinations of CNNs based on modality before combination and being fed into an LSTM \cite{chen_deep_2021}. To focus upon more important time points within the input sequence, Singh et al. \cite{singh_deep_2021} include a self-attention mechanism in a CNN-LSTM. Attention is also used with multilevel residual networks \cite{al-qaness_multi-resatt_2023} and deep triplet networks \cite{tang_triple_2022}. However, when dealing with short windows of activities an RNN component is of limited use, since there is little time dependency within a five second span of a person driving a car, for example.
\par
Given a well chosen backbone, an RNN component is not strictly necessary for SOTA results, CNN/ResNet alone can be used for classification, demonstrated by Mekruksavanich et al. \cite{mekruksavanich_resnet-se_2022}, applying a ResNet-based model directly to activity time series, while Xu et al. \cite{xu_human_2020} convert a HAR task to a typical computer vision problem by using the Gramian Angular Field algorithm to convert the 1D time series into 2D images and applied deep ResNets to the resulting images. Domain knowledge can also be leveraged, such as by using symbolic reasoning and contextual information alongside deep neural networks for prediction \cite{arrotta_semantic_2024}. Though effective, these methods are often constrained to specific data forms and collection procedures. 
\subsection{Out-of-Distribution Detection \label{section:relatedworks:ood}}

Open-set classifiers are typically set up as closed-set classifiers for classifying the ID samples with an added OOD detection component for distinguishing between ID and OOD samples which can then overrule the classification by flagging the sample as OOD. Post-hoc techniques are often the simplest to implement, since they require no adjustments to the underlying model, and can identify OOD samples by using a scoring function such as thresholding maximum softmax probability \cite{hendrycks_baseline_2018}, energy score \cite{liu_energy-based_2020}, GradNorm score \cite{huang_importance_2021}, or fitting a distribution to the activations \cite{bendale_towards_2016}. As straightforward scoring functions can have different effects depending on scenario and model setup they can complicate implementation. To improve OOD classification, the model activations themselves can also be adjusted at inference using ASH \cite{djurisic_extremely_2023}. This involves simplifying the model representation during inference, compromising ID classification. Temperature scaling can also be used for calibration, for example ODIN \cite{liang_enhancing_2020}, although this also requires perturbing inputs to further separate ID samples from OOD. Boyer et al. \cite{boyer_out--distribution_2021} evaluate many of these post hoc methods for HAR applications, and find that state-of-the-art OOD detection techniques are not effective in the HAR domain. 
\par
Instead of creating scoring functions based on model activations, training an OOD classifier on OOD samples can be a more direct approach. However, in the absence of OOD data during training (by definition), OOD samples must be simulated using the available training data. Outliers can be used as surrogate OOD examples, with output regularised to produce lower confidence on either manually collected outliers \cite{hendrycks_deep_2019, wang_out--distribution_2023}, or on virtual outliers produced by the model \cite{tao_non-parametric_2023, du_vos_2022}. Both sources of outliers have drawbacks, either from the extra labour required, or from relying on the quality of the feature space. Nie et al. \cite{nie_out--distribution_2025} circumvent this by directly perturbing input images in the pixel space, but this approach lacks a direct analogue in the HAR domain. Generative models can also be incorporated into the open-set classification framework using the assumption that generative models trained on OOD data will have more trouble reconstructing OOD samples than ID samples \cite{gao_diffguard_2023, zhou_rethinking_2022, perera_generative-discriminative_2020}. The associated reconstruction loss is used as an OOD scoring function. Tonmoy et al. \cite{tonmoy_hierarchical_2021} and Li et al. \cite{li_multiresolution_2023} both use this approach for open-set classification on HAR datasets. However, the assumption that OOD samples are harder to reconstruct than ID may not always hold \cite{nalisnick_deep_2019}, and additionally, generative models are expensive and require increased testing time in the inference stage.
\subsection{Public HAR Datasets \label{section:relatedworks:public_datasets}}

Due to the innumerable possible applications and useful scenarios for HAR, a great many datasets have been collected using assorted sensors, sizes, set ups, and purposes. A broad distinction can be drawn between vision-based and sensor-based datasets. Vision-based datasets \cite{kay_kinetics_2017, kuehne_hmdb_2011, sigurdsson_charades-ego_2018} use input of videos of subjects performing physical activities, while sensor-based datasets provide input from body worn Inertial Measurement Unit (IMU) devices, usually containing accelerometer and gyroscope sensors, and also commonly containing some selection of temperature, heart rate, magnetometer and altitude sensors, among others. 
\par
IMU devices are cheap and non-invasive, making large-scale data collection relatively feasible, and projects such as the UK-Biobank \cite{doherty_large_2017} and NHANES \cite{belcher_us_2021} have produced huge amounts of sensor data. However, this data is unlabelled, which is the primary bottleneck in HAR data collection. As such, annotated datasets are generally small and tailored to a specific purpose. UniMiB-SHAR \cite{micucci_unimib_2017} and MobiAct \cite{chatzaki_human_2017} are two popular HAR datasets, collected for training fall detection models, passively identifying when a patient might have injured themselves. Their activity classes are split between activities of daily living (ADL) like sitting or walking, and different types of falls. Other HAR datasets such as MotionSense \cite{malekzadeh_protecting_2018} were collected to investigate whether physical and demographic attributes can be inferred from sensor data, and REAL-DISP \cite{banos_benchmark_2012} looks at the impact of inconsistent sensor placement on the body. These datasets contain high quality HAR data, however the original focus of the collection process often means a significant portion must be discarded when utilising them for HAR classification. Information also must be discarded when multiple heterogenous devices are used, for example when using the dataset HHAR \cite{stisen_smart_2015}. During the acquisition of this dataset, each subject wore eight smart phones and four smart watches, but due to differences in sampling rates and sensors between devices, for most use cases only a single device can be used for training/testing. Smart devices are still useful and popular for data collection thanks to their wide availability and application relevance. Two of the most commonly studied HAR datasets, WISDM \cite{kwapisz_activity_2011} and UCI HAR \cite{anguita_public_2013}, have annotated IMU data from smart phones for six classes of ADL, although they also suffer from inconsistent sampling across devices. The six classes of ADL: standing, sitting, laying down, walking, downstairs and upstairs are the same classes in almost all the above described datasets, with datasets rarely having any more than two classes not from this set of activities. This lack of activity diversity makes OOD examination by omitting classes during training difficult, since a single class represents a large portion of total data available.
Some HAR datasets such as OPPORTUNITY do contain a NULL class alongside the defined activities, typically denoting unlabelled or unguided periods of activity. Since these periods are often inherently untracked and NULL is used to label all activities taking place during unscripted moments, it can occur that the NULL data is not truly distinct from the listed classes, which can lead to deflated or misleading results. Datasets containing typical ADL often have problematic overlap between ID and NULL classes \cite{boyer_out--distribution_2021, cherian_exploring_2024}. Therefore, while NULL classes can function as OOD sets, using omitted ID classes is preferable to ensure truly distinct sets of OOD activities. 
\par
For our analysis, we identified three datasets from the HAR community. Firstly, PAMAP \cite{reiss_introducing_2012} which is based on three IMUs on the wrist, chest, and ankle. The dataset has nine subjects, however only eight had complete data for all twelve activity classes. OPPORTUNITY \cite{chavarriaga_opportunity_2013} has four ADL from four subjects, each wearing seven IMUs on the arms, back, and legs. Participants wore twelve additional Bluetooth accelerometers, however they were not usable due to high levels of noise. Activities were annotated at different levels of specificity. Due to missing data on some subjects for higher levels of specificity, we target the locomotive level of activities, consisting of four classes of motion. Finally, RealWorld \cite{sztyler_-body_2016} is the largest of the three datasets in terms of length (18 hours). Fifteen subjects wore seven mobile devices (six smartphones and a smartwatch) located on the head, torso, arms, and legs. Subjects performed eight different activities for roughly ten minutes per activity, with the exception of jumping. For all three datasets the IMUs contained accelerometer, gyroscope, and magnetometer sensors.
\section{Method} \label{section:method}

We will first give an overview of our approach. \modelname takes as input raw sensor signals collected from one or more IMU devices which are fed into a ResNet-based feature extractor (Section \ref{section:method:feat_extractor}). The output vector is of size $|H|$, the number of nodes in the hierarchy, determined by the generation process, where $|H| > |K|$, the number of ID classes in the dataset. The hierarchy $H$ is generated using self-supervised Hierarchical Agglomerative Clustering (HAC). Leaf nodes correspond directly to classes in the dataset, and the internal nodes correspond to implicit activity categories that emerge from the hierarchical relationships between the labelled classes, where their semantic interpretation is derived from the known activities they encompass (Section \ref{section:method:hierarchy_generation}). Each class $k \in K \subseteq H$ has an associated path likelihood used to classify the sample (Section \ref{section:method:hierarchical_classification}). If along the prediction path, decisions between nodes have high enough decision entropy ($\equiv$ low prediction confidence) on average, they are labelled OOD (Section \ref{section:method:ood_score}). During inference on OOD samples, sufficiently high entropy on a node during traversal will trigger the inference stopping criterion, terminating prediction at this step and returning the parent node as the final prediction (Section \ref{section:method:inference_stopping_criterion}). This allows the model to still provide meaningful information to the user when the class cannot be determined (Section \ref{section:method:prediction_evaluation}). During training, it is therefore also necessary to temper overconfidence in predictions by simulating OOD data using only ID classes from the training set (Section \ref{section:method:loss_function}). An overview of the procedure is given in Figure \ref{fig:main_overview}.

\subsection{Feature Extractor \label{section:method:feat_extractor}}

\begin{figure}[ht]
    \centering
    \includegraphics[width=1.0\textwidth]{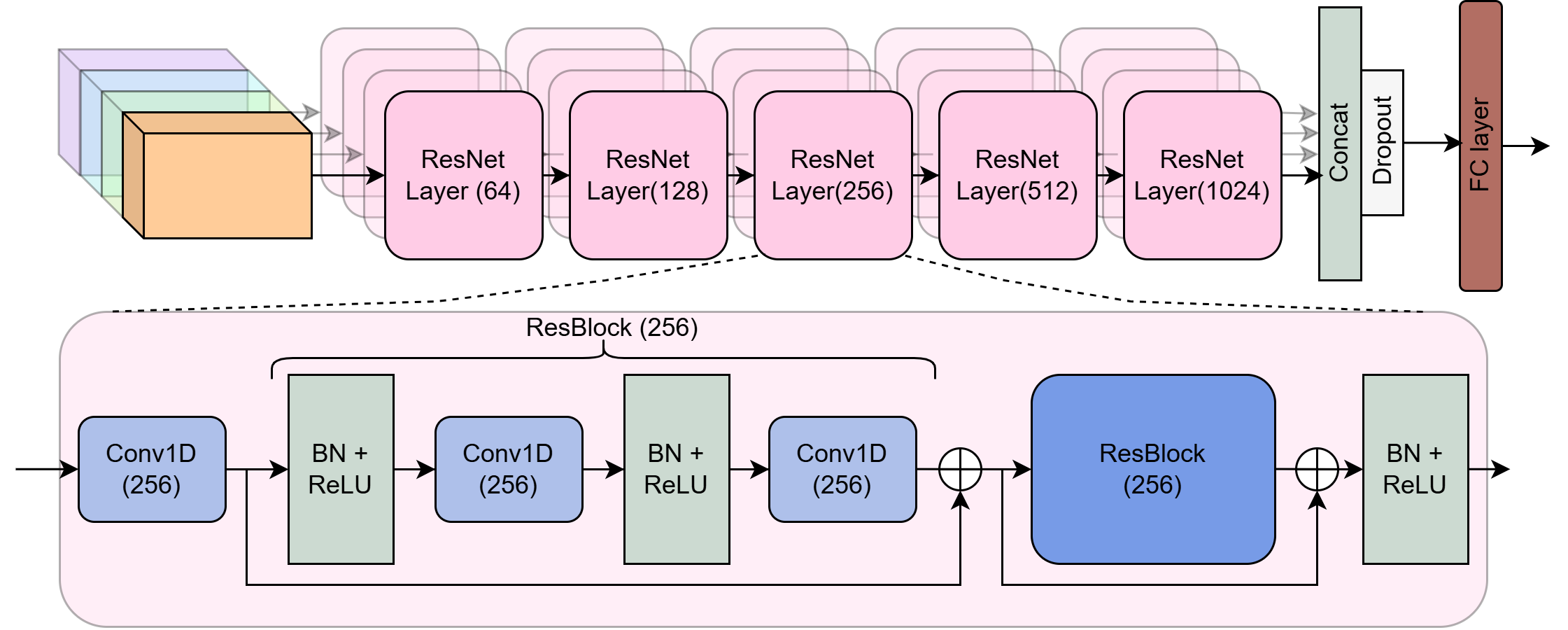}
    \caption{Feature extractor architecture adapted from Yuan et al. \cite{yuan_self-supervised_2022}, extended to the multi-sensor, multi-device domain.}
    \label{fig:feat_extractor}
\end{figure}    

The feature extractor component is purely CNN-based, as in Mekruksavanich et al. \cite{mekruksavanich_resnet-se_2022}, and is adapted from Yuan et al. \cite{yuan_self-supervised_2022}. Raw time series input is separated at the sensor level and fed through the feature extractor in parallel. As shown in Figure \ref{fig:feat_extractor}, each parallel component consists of five successive ResNet layers, increasing the number of channels up to 1024. The original architecture was designed for a single accelerometer device. The feature extractor has been duplicated along the sensor axis to permit parallel inputs from multiple devices and sensors, with input channels also adjusted to accept non-triaxial sensor input. Each parallel feature extractor is also initialised with the pretrained weights from \cite{yuan_self-supervised_2022}. The output of the final ResNet layer is concatenated to the outputs of the other parallel blocks followed by a dropout operation. A Fully Connected (FC) layer then outputs a vector of length $|H|$ containing probabilities for every node in $H$. During training, all the weights of the ResNet layers are fully trainable and fine tuned, while the fully connected layers are simultaneously trained from scratch. 
\subsection{Hierarchy Generation \label{section:method:hierarchy_generation}}
\begin{figure}[ht]
    \centering
    \centering
    \includegraphics[width=0.7\textwidth]{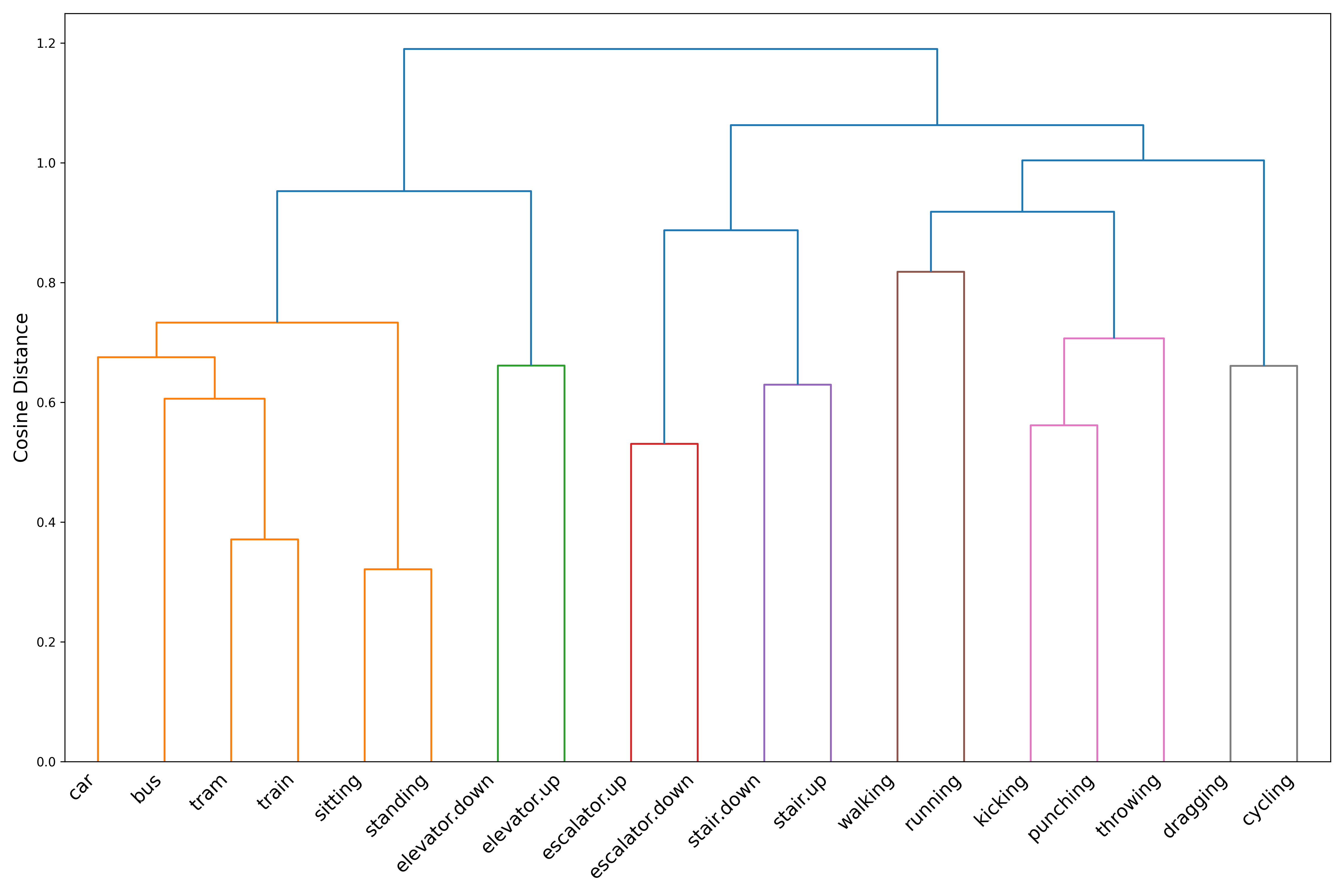}
    \caption{Dendrogram hierarchy generated using Hierarchical Agglomerative Clustering on \datasetname with total Cosine Distance after each merge indicated on the y-axis.}
    \label{fig:dendrogram_hier}
\end{figure}

Class hierarchies outside of HAR are typically constructed from domain knowledge and ontologies. No definitive hierarchy exists for HAR classes as constructing one is difficult due to a lack of objective rankings of activity classes meaning expert opinions can differ significantly. One problem specific to the HAR domain is the mismatch between semantically related activities and physically related activities. Two activities, such as \textit{standing} and \textit{elevator} can be completely distinct semantically despite being almost identical in terms of physical performance. When incorporating a hierarchy into a classification system, it is preferable to emphasise physical similarities which are present in the data over semantic ones, which are not. As such, deriving a hierarchy with unsupervised clustering such as Hierarchical Agglomerative Clustering (HAC) is better suited to HAR than using pre-defined ontologies. HAC takes a bottom up approach, each class starts in its own cluster and pairs of clusters are merged until all classes have been combined into the Root.  
\par
To create the hierarchy, firstly the feature extractor from Section \ref{section:method:feat_extractor} is trained for five epochs on the training set. The centroids of the classes in the embedding space are then used to form the initial $K$ clusters. The two closest clusters by cosine distance are merged. After merging the two closest clusters, the process is repeated until all classes are within a single cluster. The final output is a complete hierarchy of $|H|$ nodes, including  all $K$ classes as leaf nodes. An example of such a hierarchy generated from \datasetname using all available classes is shown in Figure \ref{fig:dendrogram_hier}, including the total cosine distance present after each merge.
\subsection{Loss Function \label{section:method:loss_function}}

\begin{figure}[ht]

    \centering
    \begin{subfigure}[b]{0.4\textwidth}
      \centering
      \includegraphics[width=\textwidth]{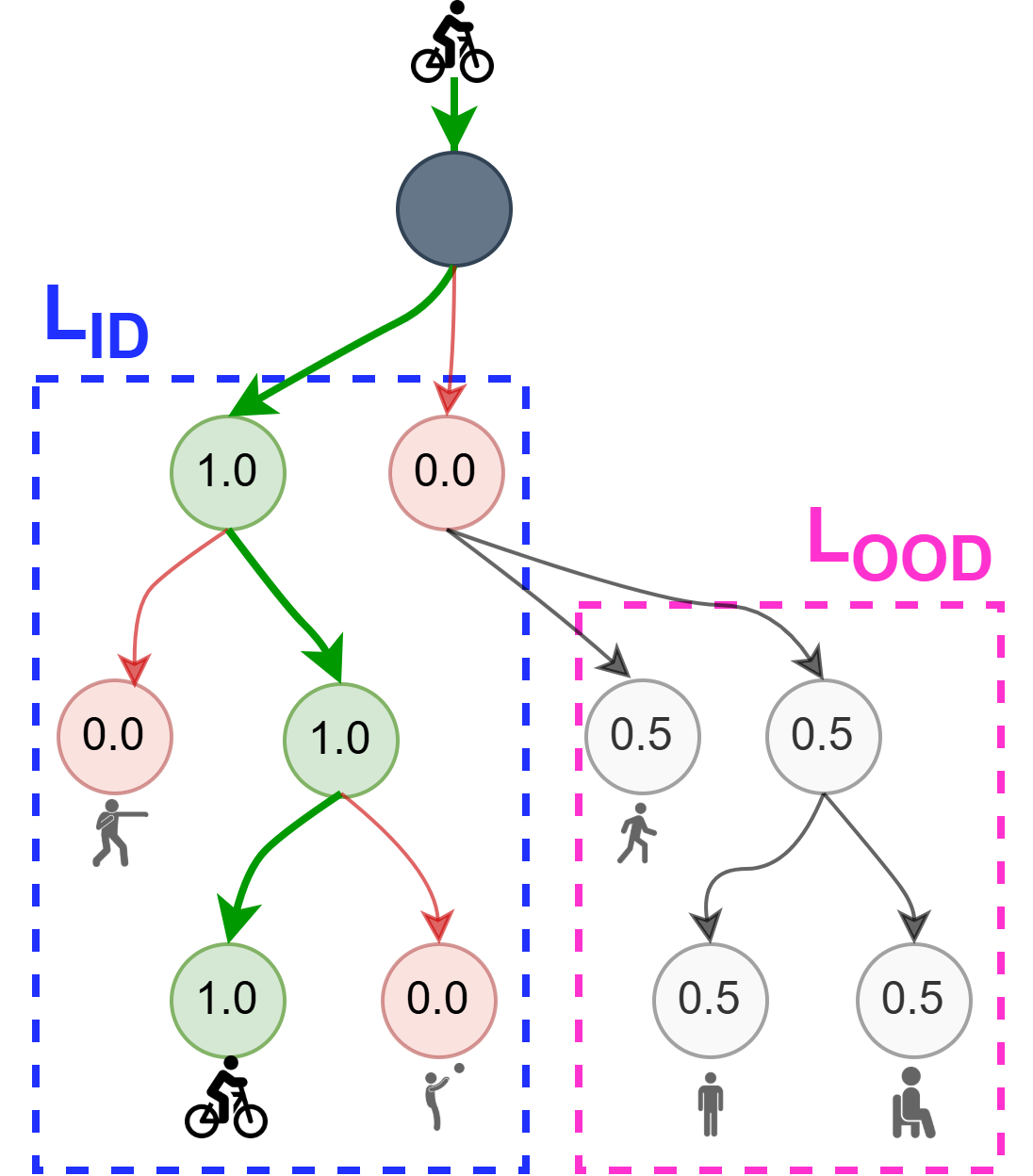} 
      \caption{Training}
      \label{subfig:loss_f_explanation}
    \end{subfigure}
    \hfill
    \begin{subfigure}[b]{0.5\textwidth}
      \centering
      \includegraphics[width=\textwidth]{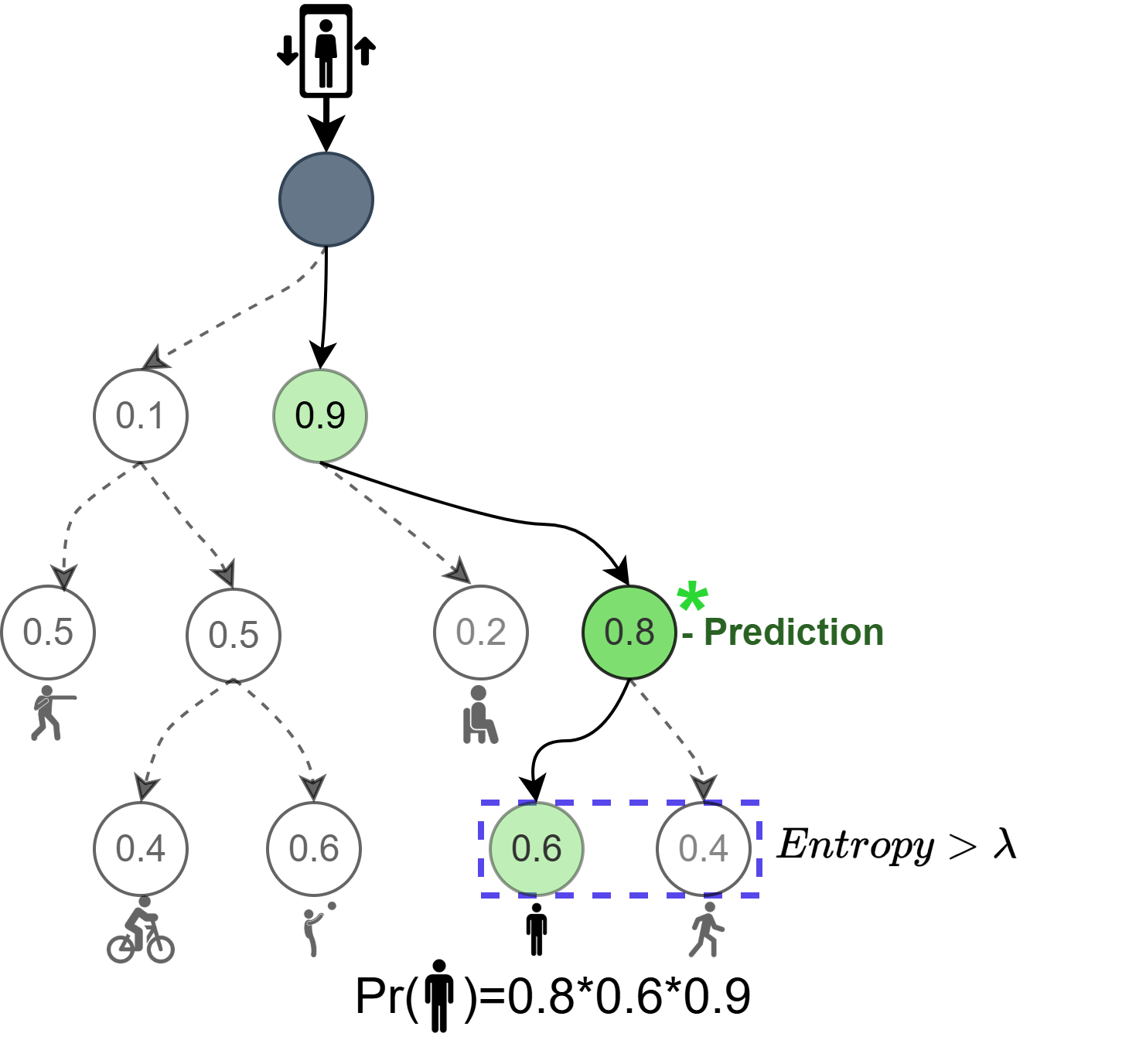} 
      \caption{Inference}
      \label{subfig:hierarchy_explanation}
    \end{subfigure}

    \caption{Illustrative example of training and inference on the Hi-OSCAR hierarchy. (a) Visualisation of optimal activations during training. Nodes along the path to correct ID class are optimised by $L_{ID}$, while nodes off the path are optimised to the uniform distribution by $L_{OOD}$ (b) Example visualisation of inference on OOD class \textit{elevator}, showing example node activations. Standing has the highest path probability, however due to high entropy in the final level, an internal node is outputted.}
    \label{fig:method_explanation}

\end{figure}    

Given Hierarchy $H = \{0,1,...,|H|-1\}$ generated by HAC, there are $|K|$ leaf nodes, corresponding to classes in the training set, and $|H|-|K|$ internal nodes. For each class node $k \in K$, there exists a unique path $anc(k)$ from the Root node to $k$ (green path in Figure \ref{subfig:loss_f_explanation}). The loss function should encourage each node $n \in anc(k)$ to have an activation of 1, and its sibling to have activation 0, visualised in the blue box in Figure \ref{subfig:loss_f_explanation} ($L_{ID}$). When nodes are neither in $anc(k)$ or a sibling of a node in $anc(k)$, the loss function should maximise entropy at this split. Given that HAC produces trees with all binary splits, this is achieved when the sibling nodes each have activation of 0.5, shown in the pink box in Figure \ref{subfig:loss_f_explanation} ($L_{OOD}$). The computation of the two loss functions, $L_{ID}$ and $L_{OOD}$, are described below.
\par
Given true class $y \in K$, $L_{ID}$ is the cumulative Cross Entropy (CE) loss of each node in $anc(y)$:
\begin{equation}
    L_{ID} = \mathbf{W}_y \cdot \sum_{n \in \{anc(y), y\}} CE[p(n),\mathbf{y}] 
    \label{eqn:l_id}
\end{equation}
where $\mathbf{W}_y$ is the inverse frequency of class $y$. This weighting term helps account for class imbalances within the training set.
\par
The role of $L_{OOD}$ is to maximise entropy (minimise confidence) of any split in the hierarchy not containing a node in $anc(y)$. This uses ID samples as pseudo-OOD for all nodes not intersecting with $anc(y)$. These nodes $n \notin anc(y)$ are optimised to the Uniform distribution $\mathbf{U}$ using Kullback-Leibler divergence:
\begin{equation}
    L_{OOD} = \sum_{n \notin anc(y)} KL[p(n), \mathbf{U}].
\end{equation}
\par 
The two losses are then combined to produce the final loss function:
\begin{equation}
    L = L_{ID} + L_{OOD}
\end{equation}
\subsection{Hierarchical Classification \label{section:method:hierarchical_classification}}

With Hierarchy $H = \{0,1,...,|H|-1\}$ generated by HAC, each of the $|H|$ output nodes of the model are assigned to a node, either internal or class. Softmax is applied to each pair of sibling nodes, so that for any given parent node, the activations of its children are normalised and sum to one. For each class $k \in K$, there exists a unique path from root node to class node $anc(k)$. This path likelihood is calculated by multiplying the softmax score of each node along the path $anc(k)$:
\begin{equation}
    Pr(k) = p(k) \, \cdot \prod_{n \in anc(k)} p(n)
\end{equation}
where $p(n)$ is the softmax score of node $n$. During inference, the path likelihood of all classes is calculated and the provisional predicted class is assigned the class with maximum value: $\hat{k} = max_{k \in K} Pr(k)$. Calculating total path likelihood for every class $k$ rather than greedily choosing the maximum score at each node allows for mistakes earlier in the hierarchy to be corrected downstream \cite{wan_nbdt_2021}. Example path likelihood calculation is shown in Figure \ref{subfig:hierarchy_explanation}.
\subsection{OOD score \label{section:method:ood_score}}

Once the provisional predicted class $\hat{k}$ has been chosen, the path $anc(\hat{k})$ is then traversed to determine if the sample is OOD. For this, we employ the entropy of each decision along the path $anc(\hat{k})$. At each node $n \in anc(\hat{k})$, the entropy of its children is calculated:
\begin{equation}
    \mathbf{H}(n) = \sum_{n' \in ch(n)} -p(n')\text{log}(p(n')).
    \label{eqn:decision_entropy}
\end{equation}
We utilise mean path entropy $\mathbf{H}_{mean}$ as the OOD score, calculated as:
\begin{equation}
    \mathbf{H}_{mean}(\hat{k}) = \frac{1}{len(anc(\hat{k}))} \sum_{n \in anc(\hat{k})} \mathbf{H}(n).   
\end{equation}
The higher the value of $\mathbf{H}_{mean}(\hat{k})$, the less confidence there is in the prediction, and the more likely it is to be OOD.  

\subsection{Inference Stopping Criterion \label{section:method:inference_stopping_criterion}}

Beyond a simple accept/reject decision for OOD samples, Hi-OSCAR returns informative labels in the form of internal node predictions, providing richer OOD classifications which embed information about the most closely related classes. For every ID sample, the entropy $\mathbf{H}(n)$ of each node pairing $n \in anc(\hat{k})$ is calculated and stored. A threshold for stopping inference ($\lambda$) is chosen based on the values within the training set, e.g. choosing the 99\textsuperscript{th} percentile would result in inference stopping whenever a node's entropy is higher than 99\% of the entropies recorded by that node in the ID data. We denote this by $\hat{\lambda} = 0.99$. Figure \ref{subfig:hierarchy_explanation} illustrates an example of the inference stopping criterion being exceeded and outputting an internal node for an OOD sample.
\par
The choice of $\hat{\lambda}$ depends on the chosen application's fidelity requirements. Higher values of $\hat{\lambda}$ will be more prone to 'over-prediction', i.e. terminating at a leaf node or deep in the tree, while lower values of $\hat{\lambda}$ will give coarser, less precise predictions of nodes closer to the Root. 
\par
The returned node for an OOD sample should be close to the leaf nodes of similar activities in the training set, informing the user about its dynamics, and the closer the predicted internal node is to a leaf nodes (measured by cosine distance), the more similar it is. Conversely, if the OOD sample is completely novel relative to the training set, the internal node predicted will be further from the leaf nodes, and closer to the Root of the hierarchy.
\subsection{Prediction Evaluation \label{section:method:prediction_evaluation}}

Interpretation of the internal node OOD predictions requires a measure of distance to determine how 'close' or 'far' the prediction is from the ID classes. The generated hierarchy $H$ can be used for this purpose, since it already has grouped the classes and internal nodes based on similarity. Hierarchical distance, counting the number of edges between two nodes, can be used in this case. However, it assigns equal value to all edges, ignoring the differences in cosine distance associated with each merge. This can lead to inflated hierarchical distances for some classes when many similar activities are present in the dataset. Consider in Figure \ref{fig:dendrogram_hier}, \textit{train} and \textit{car} are separated by four edges, the same as \textit{throwing} and \textit{walking}, despite the former pair being closer in terms of cosine distance on the Y-axis.
\par
We therefore propose weighting each edge by the increase in total cosine distance added by that merge during the HAC process, terming it Cumulative Cosine Distance. Given predicted node $s$ and target node $t$ with $lca(s,t) = q$, Cumulative Cosine Distance is calculated as:
\begin{equation}
    D(s,t) = \sum_{\substack{u \supset s \\ u \subset q}} d(u,s) + \sum_{\substack{v \supset t \\ v \subset q}} d(v,t),
    \label{equation:cum_cos_distance}
\end{equation}
where $d(u,v)$ is the cosine distance. Lower values indicate the nodes are close i.e. similar in nature. 

\section{Dataset \datasetname} \label{section:our_dataset}

\begin{figure}[ht]
    \centering
    \includegraphics[width=0.75\textwidth]{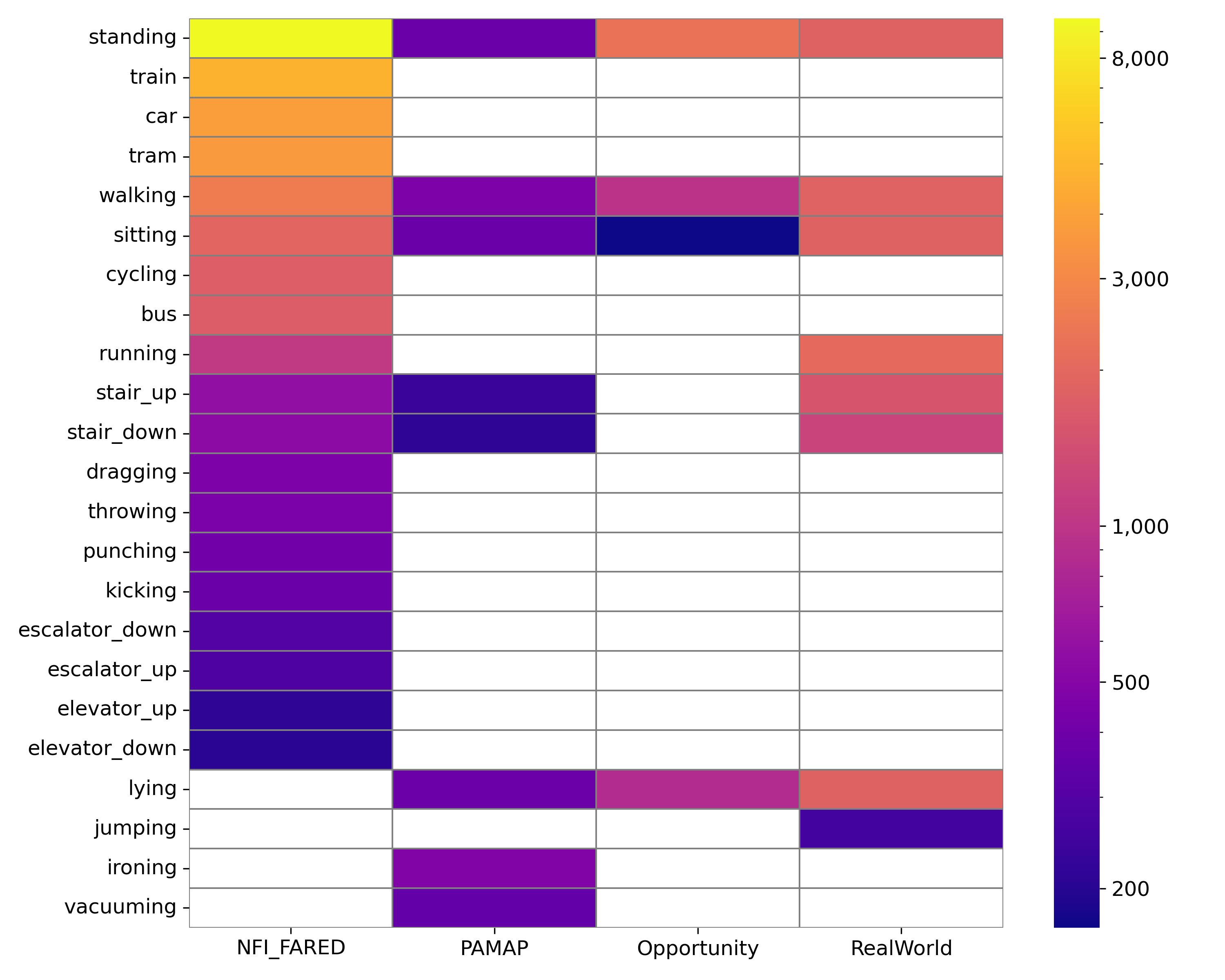}
    \caption{Number of 10 second samples from each dataset. Blank squares denote the activities not in that dataset.}
    \label{fig:hmat_dataset_activities}
\end{figure}
\begin{table}
\begin{center}
\begin{tabular}{ c|c|c|c|c|c } 
 \hline
 Model & \#Subjects & \#IMUs & \#Activities &  \#Hours & Sensors    \\
 \hline
 PAMAP      & 8  & 3 & 8 & 11 & accelerometer, gyroscope, magnetometer   \\ 
 Opportunity& 4  & 5 & 4 & 7  & accelerometer, gyroscope, magnetometer   \\
 RealWorld  & 15 & 5 & 8 & 18 & accelerometer, gyroscope, magnetometer   \\
 \datasetname   & 11 & 2 & 19 & 40 & accelerometer, gyroscope, altimeter   \\ 

 \hline
\end{tabular}
\caption{Details of datasets used in analysis}
\label{tab:dataset_summaries}
\end{center}
\end{table}
The \datasetname dataset contains data from eleven participants wearing multiple devices and sensors. The dataset contains a diverse range of activities roughly grouped as everyday movements, elevation change, dynamic movements, and transportation, totalling 19 activities for each subject. The dataset consists of a large amount of data, totalling 39 hours 48 minutes (14,267,742 samples). The selection of activities, as well as their abundance relative to other publicly available HAR datasets, is shown in Figure \ref{fig:hmat_dataset_activities}. 
\par
Below we detail the procedure followed to collect the \datasetname, as well as details on pre-processing.  

\subsection{Participants}
Fourteen Participants are included (age 26.6 $\pm$ 8.8y), made up of 6 females and 8 males. Each subject signed an informed consent form, agreeing to the collection and spreading of their data anonymously. Height and weight data of each subject cannot be shared for privacy reasons. For each subject, data was collected during multiple sessions spread over several days. Due to scheduling conflicts, three participants could not attend all sessions, and three different subjects completed those sessions in their place. Their data was combined, providing effectively eleven participants worth of data.

\subsection{Sensors}
The two deployed IMUs measure linear acceleration with triaxial accelerometer (freq = 100 Hz), angular velocities with a triaxial gyroscope (freq = 100 Hz), air pressure with an altimeter (freq = 1 Hz), and temperature (freq = 1 Hz). The inclusion of air pressure data, a rarely utilised sensor in this domain, is notable. Air pressure is naturally informative for the many activities which include elevation changes. It is also a one dimensional signal, unlike acceleration and rotation, which occur in three dimensions. During analysis, the temperature signal was omitted since it is not related to body kinematics. This resulted in thirteen channels of data used in the analysis.
\par
The IMU devices were placed on the lower back and on the dominant forearm of the subjects. The participants attached the devices themselves to introduce some heterogeneity in sensor placement. 

\subsection{Protocol}
The dataset contains 19 activities, listed in Figure \ref{fig:hmat_dataset_activities}. The activities were recorded in four separate measurement sessions. Session one includes everyday activities, session two recorded elevation change activities, session three recorded dynamic activities, and session four recorded transport activities. The sequence of performed activities within each session was randomised for each subject.
\par
Sessions one, two, and three were performed under controlled conditions. In between each activity the subject would rest one minute by either sitting or standing, as instructed by the instructor, who recorded the activities. It was found that sitting on a chair with a backrest influenced the accelerations recorded by the IMU on the back, so sitting breaks of both kinds (with and without backrest) were recorded for each subject. To encourage diversity in their execution, while performing the activities, the subjects were given as few instructions as possible. At the end of each session, the subject completed a "free-living" section, in which they would perform the activities from that session in any order and for however long they wished without breaks. For both of these scenarios, the start and end times of each activity would be recorded. Here we describe the data collection process for each session:

\paragraph{Everyday Movements Session}
The movements walking, running and cycling were  performed multiple times for known distances. For running and walking there are two distances: $\pm$240m and $\pm$90m, each performed twice. For cycling, the subjects cycled a set number of loop around the car park of the experimental location three times. The number of loops each were varied between two, three, and four. During the static sitting movement, the participant sat on a bench without backrest. 

\paragraph{Elevation Change Session}
The movements stair, escalator and elevator were performed four times each going upstairs and downstairs. The stair climbing and escalator movements were performed at different velocities. For stair climbing the subject was instructed to walk or to run the stairs. For the escalator, the subject was instructed to stand or walk on the escalator. After each dynamic movement the participant would either stand or sit for one minute at the end of the stairs/escalator. During the static sitting movement, they sat on a bench with backrest. 

\paragraph{Dynamic Movements Session}
The dynamic movements: kicking, throwing, punching and dragging were done four times for different time intervals: 10 seconds, 20 seconds and two times 30 seconds. During the kicking and punching movement, the participant would kick or punch, respectively, a punching bag. The kicking/punching was alternated between their dominant and non-dominant leg/arm. For the throwing task, the subject threw a ball using their dominant arm down a hallway to the another person, who rolled it back to the subject to pick it up and throw again until the time was complete. The ball weighed 1 kg. For the dragging activity the subject dragged the punching bag weighing 34 kg through a hallway. 

\paragraph{Transportation Session}
In session four, the subjects wore both IMUs while travelling home. The transport modes of riding the train, tram, and in a car were being recorded. If the subject had not used one or more of the listed modes of transport in their trip home, arrangements were made to ensure data for all transport types were recorded for each individual e.g. if they had not travelled by bus or car, they were then driven to the bus station, from which they would ride a bus back to an agreed location to return the IMUs. The participants recorded the start and end times of each travel activity by themselves.

\subsection{Data acquisition}
The starting times of the measurement for both devices were synchronized. The raw signal data from the IMUs were pre-processed using proprietary scripts. The data was labelled by matching the timestamps of activities recorded by the instructor with the timestamps of the raw signals. During the transport session, the data was self-reported by the subject making a note of the start and end time of each transport type. Where there was misalignment in the labels as a result of the self-labelling, e.g. a subject's watch being set wrong, labels were adjusted by a minute or two, using the signals of the sensors to find the actual transition moment between activities. It was found to be possible to locate these transition moments by eye from the sensor data, with clear indications of activity changes that aligned with the duration of activity reported by the subject, only with the start and end times shifted. Given the amount of transport data collected, this adjustment only affected a small minority of the total data. Accelerometer, gyroscope, and air pressure data are included in the dataset. The air pressure data was linearly interpolated to the frequency of 100Hz in order to match the frequency of the other signals. The data is fully publicly available for download, and scripts for processing data into windows are provided in the project GitHub. 
\section{Experiments} \label{section:experiments}
To assess open-set classification performance we test on both ID and OOD tasks (Sections \ref{section:experiment:activity_classification} and \ref{section:experiment:ood_detection}, respectively), comparing against state-of-the-art baselines from HAR specific methods and generic Machine Learning methods. In Section \ref{section:experiments:ood_detection_different}, we investigate the differences in performance for different OOD classes, and in Section \ref{section:experiment:threshold_ward_d} the impact of threshold $\hat{\lambda}$ on prediction quality. Section \ref{section:experiment:diff_hierarchies} analyses the effect of different hierarchy generation methods, both data- and knowledge-driven, and Section \ref{section:experiment:window_size} looks at the influence of window size upon performance.
\par
All experiments were performed using k-fold cross-validation on the subject level. Datasets \datasetname
and RealWorld had two subjects in each fold, while OPPORTUNITY and PAMAP had one, due to the number of subjects available in the datasets. Data was split subject-wise to prevent data leakage and to better evaluate generalisability to new subjects' data. The hierarchy used by Hi-OSCAR was always generated using only samples from the training set to further prevent data leakage. Each fold was then repeated five times, reporting the mean and standard deviation (in brackets). The data was divided into 10s windows with 50\% overlap and downsampled to 30 Hz. Training was run for a maximum of 250 epochs using early stopping, which terminated training if validation loss did not decrease for 5 epochs in a row, to prevent overfitting. 
\par
\modelname was trained with the Adam optimiser with learning rate 1e-4, betas = (0.9,0.999). All experiments were performed using an NVIDIA RTX 4500 Ada Generation GPU. Baseline models used for comparison were replicated from the original authors' descriptions and code (if available) and implemented in PyTorch. For baselines not taken from literature or when hyperparameters were not specified, hyperparameter tuning was used to set values. Further details are provided in Appendix \ref{section:appendix:hyperparameters}. To enable fair and meaningful comparisons between models, we standardised the evaluation protocol (cross-validation, window size, dataset choice, etc.). As a consequence of standardisation, there are some differences in performance between the reported results in the original papers and those in our experiments. For example, the popular LIMU-GRU model \cite{xu_limu-bert_2021} uses ~95\% of the training data for unsupervised learning to generate embeddings, and the remaining ~5\% for supervised training. In our set up, all baselines utilise 100\% of the data, making fair comparison with LIMU-GRU not possible. Instead we include the GRU baseline, which is comparable to the classifier used in LIMU-GRU when pretraining is not performed beforehand. 
\par

\subsection{Activity Classification \label{section:experiment:activity_classification}}
\begin{table}
\begin{center}
    
\begin{tabular}{c|c|c|c|c}
\hline
 & PAMAP  & OPPORTUNITY &  RealWorld  & \datasetname  \\
\hline
Model & F1 $\uparrow$ & F1 $\uparrow$ & F1 $\uparrow$ & F1 $\uparrow$ \\
\hline
GRU & 0.366 (0.081)  & 0.422 (0.051) & 0.384 (0.036) & 0.169 (0.025) \\
DCNN \cite{chen_deep_2015} & 0.609 (0.083) & 0.257 (0.102) & 0.226 (0.165) & 0.223 (0.088) \\
Transformer \cite{dirgova_luptakova_wearable_2022} & 0.456 (0.131) & 0.579 (0.057) & 0.597 (0.111) & 0.235 (0.056) \\
DeepSense \cite{yao_deepsense_2017} & 0.657 (0.069) & 0.551 (0.110) & 0.663 (0.073) & 0.334 (0.024) \\
ResNet-SE \cite{mekruksavanich_resnet-se_2022} & 0.534 (0.077) & 0.527 (0.067) & 0.647 (0.092) & 0.442 (0.069) \\
CNNLSTM-Att \cite{murahari_attention_2018} & \underline{0.428} (0.342) & 0.580 (0.113) & 0.586 (0.150) & 0.484 (0.101) \\
MLP & 0.759 (0.094) & 0.727 (0.031) & 0.760 (0.059) & 0.502 (0.048) \\
CNNBiGRU \cite{mekruksavanich_deep_2021} & \underline{0.823} (0.140) & \underline{0.761} (0.121) & 0.772 (0.071) & 0.584 (0.155) \\
Random Forest & \underline{0.911} (0.036) & \underline{0.820} (0.038) & \underline{0.889} (0.051) & 0.677 (0.043) \\
\modelname & \textbf{0.927} (0.055) & \textbf{0.827} (0.027) & \textbf{0.903} (0.028) & \textbf{0.827} (0.047) \\
\hline
\end{tabular}
\caption{F1 scores for ID classification on HAR datasets. Standard deviation is shown in brackets and best performing model on each dataset is in bold. Underlined scores indicate models whose performance is not statistically significantly different (p $\geq$ 0.05) from the
best-performing model, according to a paired t-test over the folds.\label{tab:id_results}}
\end{center}
\end{table}

We evaluate ID classification using \datasetname and the three public HAR datasets: PAMAP, OPPORTUNITY, and RealWorld. This first experiment looks at pure ID performance when all classes are present during training. 
\par
Performance is measured using the following metric: 
\begin{itemize}
    \item \textbf{Macro F1-score} is calculated by finding the F1-score $F1 = \frac{2 \times TP_{ID}}{2 \times TP_{ID} + FP_{ID} + FN_{ID}}$ for each class and computing the unweighted mean. Macro-F1 score assigns equal importance to all classes. An ID True Positive ($TP_{ID}$) is an ID sample correctly labelled and other quantities are similarly defined.
\end{itemize}
\par
Results of the ID experiments are shown in Table \ref{tab:id_results}. \modelname performs best on all four datasets, showing a 1.76\%, 0.85\%, and 1.57\% improvement compared to baselines on datasets PAMAP, OPPORTUNITY, and RealWorld, respectively. Significance was assessed using paired t-tests over the folds. They show that although worse on average, Random Forest did not perform significantly worse than \modelname on the three public HAR datasets. CNNBiGRU and CNNLSTM-Att also showed no significant difference to \modelname on PAMAP and OPPORTUNITY, however this can also be attributed to the high variance in their outputs, supported by a large effect size (Cohen's distance) relative to \modelname of -1.32 and -1.2 on PAMAP for CNNLSTM-Att and CNNBiGRU, respectively, compared to -0.8 for Random Forest. Similarly, on OPPORTUNITY, CNNBiGRU has an effect size of -0.7 compared to Random Forest's -0.2. 
\par
On \datasetname, \modelname beats baselines by over 20\%. All baselines struggled with the more complex task of classifying 19 activities. Random Forest, the best baseline, saw a 17\% drop in score between \datasetname and OPPORTUNITY, its second worst dataset. Meanwhile, \modelname scored the same as it did on the OPPORTUNITY dataset. 
\subsection{Out-of-Distribution Detection \label{section:experiment:ood_detection}}
\begin{table}[ht]
\centering
\begin{tabular}{c|c|c|c|c|c}
\hline 
\multicolumn{3}{c|}{} & \multicolumn{2}{c|}{OOD} & \multicolumn{1}{c}{ID} \\
\hline
Model & Dataset & OOD class & AUROC $\uparrow$ & Detection Error $\downarrow$ & F1 $\uparrow$  \\
\hline

\multirow{4}{*}{OpenNet \cite{hassen_learning_2020}} 
  & \multirow{3}{*}{\datasetname} & dragging & 0.631 (0.122) & 0.343 (0.089) & 0.734 (0.019) \\
  & & tram & \textbf{0.695} (0.097) & \textbf{0.339} (0.070) & 0.759 (0.057) \\
  & & running & \textbf{0.731} (0.184) & \textbf{0.234} (0.123) & 0.685 (0.061) \\
  \cline{2-6} 
  & OPPORTUNITY & NULL & 0.538 (0.179) & \underline{0.337} (0.120) & \underline{0.796} (0.023) \\
\hline 
\multirow{4}{*}{VAE} 
  & \multirow{3}{*}{\datasetname}& dragging & 0.471 (0.119) & 0.405 (0.034) & 0.593 (0.042) \\
  & & tram & \underline{0.506} (0.144) & \underline{0.410} (0.068) & 0.599 (0.046) \\
  & & running & 0.545 (0.164) & 0.317 (0.072) & 0.604 (0.047) \\
  \cline{2-6}
  & OPPORTUNITY & NULL & 0.451 (0.150) & 0.380 (0.051) & 0.505 (0.087) \\
\hline
\multirow{4}{*}{kNN} 
  & \multirow{3}{*}{\datasetname}& dragging & 0.693 (0.131) & 0.246 (0.022) & \underline{0.813} (0.060) \\
  & & tram & 0.465 (0.068) & 0.445 (0.044) & \textbf{0.844} (0.049) \\
  & & running & \underline{0.614} (0.127) & \underline{0.293} (0.087) & \textbf{0.816} (0.064) \\
  \cline{2-6}
  & OPPORTUNITY & NULL & 0.456 (0.177) & 0.406 (0.082) & \underline{0.838} (0.037) \\
\hline
\multirow{4}{*}{\modelname} 
  & \multirow{3}{*}{\datasetname}& dragging & \textbf{0.898} (0.067) & \textbf{0.159} (0.076) & \textbf{0.815} (0.051) \\
  & & tram & \underline{0.659} (0.096) & \underline{0.353} (0.054) & \underline{0.836} (0.023) \\
  & & running & 0.295 (0.144) & 0.460 (0.028) & \textbf{0.824} (0.050) \\
  \cline{2-6}
  & OPPORTUNITY & NULL & \textbf{0.820} (0.048) & \textbf{0.188} (0.042) & \underline{0.828} (0.050) \\

\hline
\end{tabular}
\caption{Comparison with open-set baselines, using OOD metrics AUROC and Detection Error and ID metric F1. Models are trained and tested using classes \textit{dragging}, \textit{running}, and \textit{tram} from \datasetname. F1 is calculated based on remaining ID classes. Best results are in bold. Underlined scores indicate models whose performance is not statistically significantly different (p $\geq$ 0.05) from the best-performing model, according to a paired t-test over the folds.}
\label{tab:ood_results}
\end{table}

Open-set classification was evaluated using the datasets \datasetname and OPPORTUNITY. For \datasetname, one class (the OOD class) was omitted from the training set before training under the same conditions as in Section \ref{section:experiment:activity_classification}. For OPPORTUNITY, the provided NULL class was used as OOD, and also not included in the training set. In both cases, the test set included all classes, including the OOD class. An OOD score was calculated for every sample in the test set, both ID and OOD. This OOD score is used to distinguish samples of ID classes from samples from the OOD class, where a larger OOD score means a sample is more likely OOD. Choice of OOD score depends on method. In the context of OOD detection, a True Positive ($TP_{OOD}$) is defined as a correctly labelled OOD sample, a False Positive ($FP_{OOD}$) is an ID sample incorrectly labelled OOD, and so on for the other quantities. How effective the OOD score was at correctly labelling OOD samples was measured using the following metrics:
\begin{itemize}
    \item \textbf{AUROC} Is the Area Under the Receiver Operating Curve. The Receiver Operating Characteristic (ROC) curve plots the relationship between TPR and FPR. The area under the ROC curve can be interpreted as the probability that a positive example (ID) will have a higher detection score than a negative example (OOD). Consequently, a random positive example detector corresponds to a 0.5 AUROC, and a ``perfect''
    classifier corresponds to 1.
    \item \textbf{Detection Error} measures the misclassification probability when TPR is 95\%. The definition of Detection Error is given by Detection Error = $ 0.5(FPR_{OOD} + FNR_{OOD})$.
\end{itemize}
\par
We compare \modelname to a number of baseline open-set classification methods, namely OpenNet \cite{hassen_learning_2020}, k-Nearest-Neighbours (kNN), and Variational Auto Encoder (VAE). All methods use the same feature extractor specified in Section \ref{section:method:feat_extractor} and otherwise use the training procedure required by the method, e.g. a different loss function for OpenNet. Varying only the OOD detection method is done to evaluate the contribution of the hierarchical structure to OOD performance. Such changes to the model or training routine can also impact ID performance. We therefore also include the ID metric macro-F1 score in the results (Table \ref{tab:ood_results}). Though the focus of these experiments is on OOD performance, it is important to see how it balances with ID results for reliable open-set classification.
\par
Experiments are run using three different choices of OOD class, namely \textit{dragging}, \textit{tram}, and \textit{running}. These classes cover the range of \modelname's performance and give a fuller picture of OOD ability (see Section \ref{section:experiments:ood_detection_different}).
\par
When \textit{dragging} is the OOD class, \modelname performs best, beating the best baselines kNN by 30\% on AUROC 35\% on detection error. For \textit{tram}, OpenNet is best, scoring 5\% higher than \modelname on AUROC and 4\% higher on detection error. According to paired t-tests, OpenNet and \modelname were statistically equivalent for OOD class \textit{tram}. On \textit{running}, \modelname scores worse than the baselines with AUROC 0.295, far below the best baseline OpenNet. For the NULL class from OPPORTUNITY, \modelname scores higher than all the baselines on both AUROC and detection error, beating next best model OpenNet by 52\% and 44\%, respectively.
\par
ID performance varied across the baselines, and is reduced for both OpenNet and VAE, whose training procedures negatively impact performance. kNN, the least invasive method, is statistically equivalent to \modelname on F1 score. While OpenNet and kNN match \modelname's performance on ID and OOD tasks separately, only \modelname is competitive on both dimensions.
\par
For all methods, OOD performance varies significantly with choice of OOD class. Each of the three examined OOD classes proves most difficult (measured by AUROC) for at least one of the methods. Similarly, the easiest to flag OOD class varies depending on method. Experiments using OOD sets containing multiple randomly selected classes in the OOD sets are included in Appendix \ref{section:appendix:big_ood_set}.
\subsection{Effect of Activity on Out-of-Distribution Performance \label{section:experiments:ood_detection_different}}
\begin{table}[ht]
\centering
\begin{tabular}{c|c|c|c}
\hline
& \multicolumn{2}{c|}{OOD} & \multicolumn{1}{c}{ID} \\
\hline
Excluded Class & AUROC $\uparrow$ & Detection Error $\downarrow$  & F1 $\uparrow$ \\ 
\hline
dragging & 0.898 (0.067) & 0.159 (0.076) & 0.815 (0.051) \\
cycling & 0.887 (0.063) & 0.184 (0.068) & 0.820 (0.036) \\
standing & 0.865 (0.051) & 0.207 (0.052) & 0.834 (0.038) \\
throwing & 0.783 (0.085) & 0.250 (0.059) & 0.812 (0.049) \\
sitting & 0.777 (0.088) & 0.253 (0.100) & 0.844 (0.038) \\
train & 0.761 (0.086) & 0.286 (0.046) & 0.845 (0.032) \\
escalator.down & 0.736 (0.088) & 0.297 (0.057) & 0.838 (0.045) \\
bus & 0.735 (0.099) & 0.308 (0.067) & 0.835 (0.060) \\
tram & 0.659 (0.096) & 0.353 (0.054) & 0.836 (0.023)  \\
stair.up & 0.641 (0.129) & 0.364 (0.086) & 0.825 (0.050) \\
elevator.down & 0.617 (0.059) & 0.358 (0.046) & 0.834 (0.028) \\
car & 0.597 (0.144) & 0.389 (0.061) & 0.827 (0.049) \\
kicking & 0.591 (0.304) & 0.250 (0.132) & 0.840 (0.032) \\
elevator.up & 0.572 (0.094) & 0.376 (0.023) & 0.838 (0.041) \\
escalator.up & 0.552 (0.124) & 0.396 (0.073) & 0.839 (0.024) \\
stair.down & 0.546 (0.238) & 0.383 (0.114) & 0.813 (0.039) \\
walking & 0.524 (0.119) & 0.384 (0.072) & 0.853 (0.026) \\
punching & 0.503 (0.171) & 0.406 (0.085) & 0.818 (0.027) \\
running & 0.295 (0.144) & 0.460 (0.028) & 0.824 (0.050) \\
\hline
\end{tabular}
\caption{Open-Set classification results for different selections of OOD class.\label{tab:ood_results_ours}}

\end{table}
\begin{figure}[ht]
    \centering
    \includegraphics[width=0.75\textwidth]{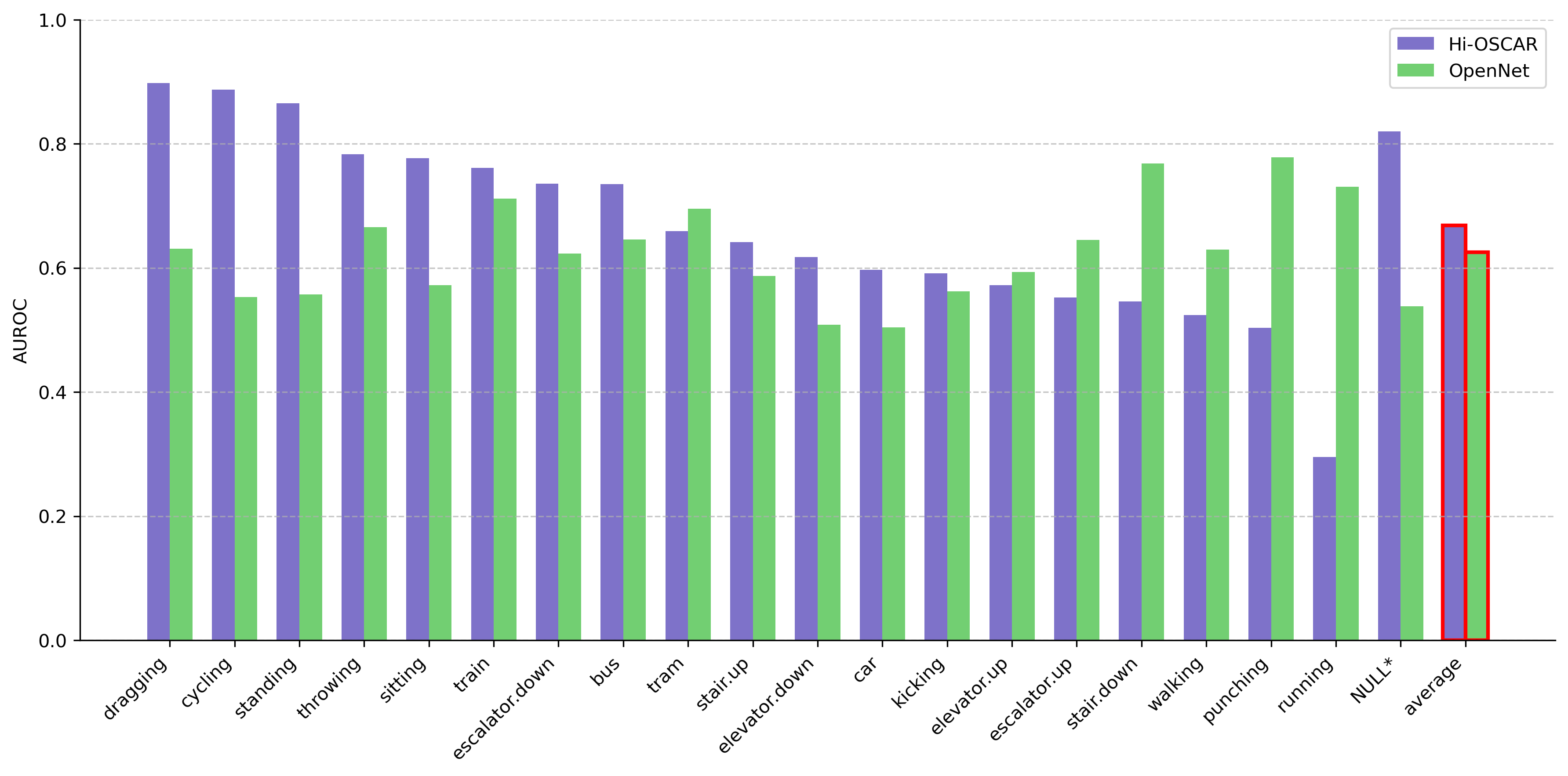}
    \caption{Comparison of AUROC between \modelname and OpenNet for different OOD classes from dataset \datasetname. *NULL class is from the OPPORTUNITY dataset}
    \label{fig:auroc_comp_hist}
\end{figure}
\begin{figure}[ht]
  \centering
  \begin{subfigure}[b]{0.45\textwidth}
    \centering
    \includegraphics[width=\textwidth]{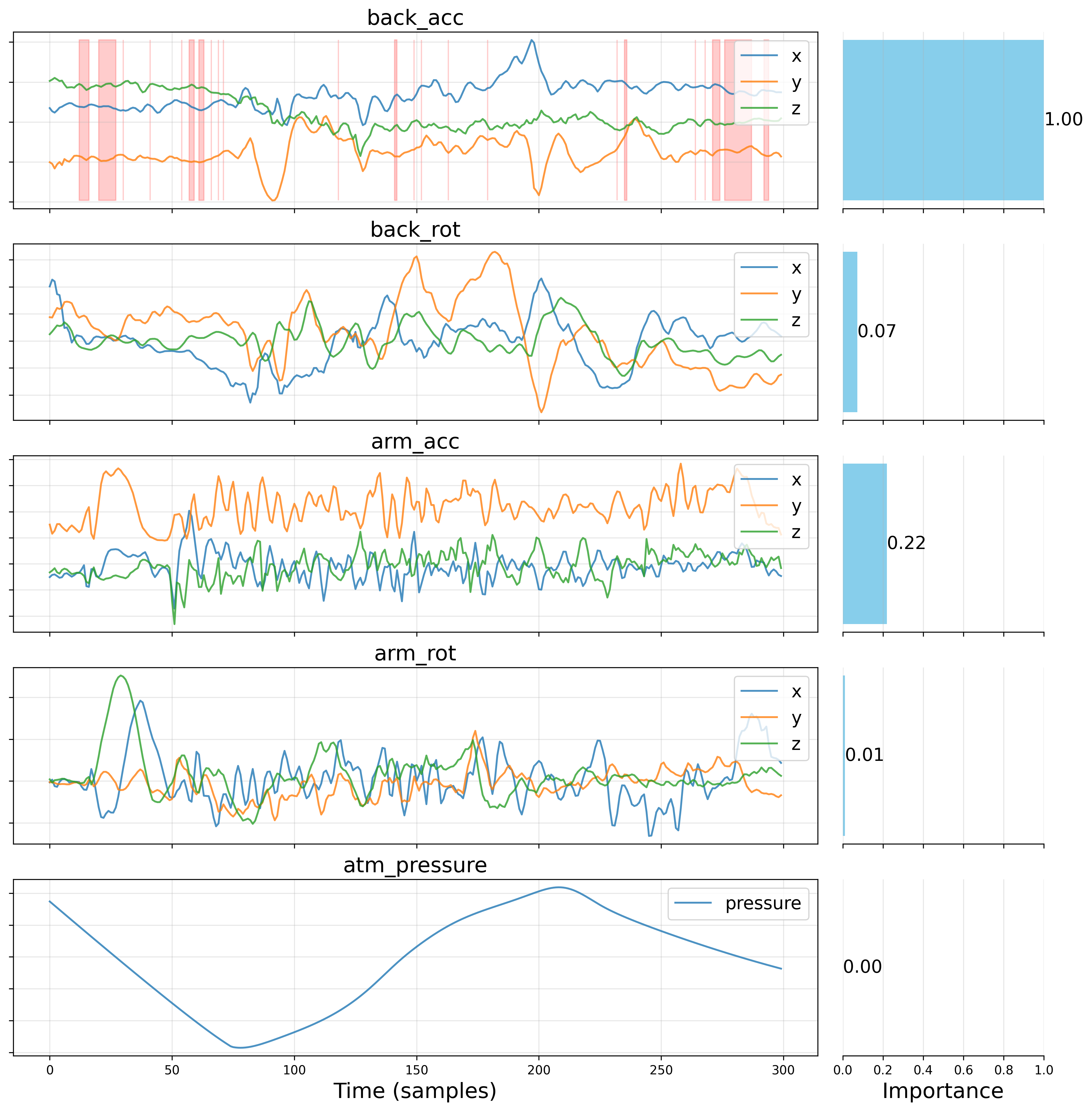} 
    \caption{\modelname: \textit{cycling}}
    \label{subfig:ours_cycling}
  \end{subfigure}
  \hfill
  \begin{subfigure}[b]{0.45\textwidth}
    \centering
    \includegraphics[width=\textwidth]{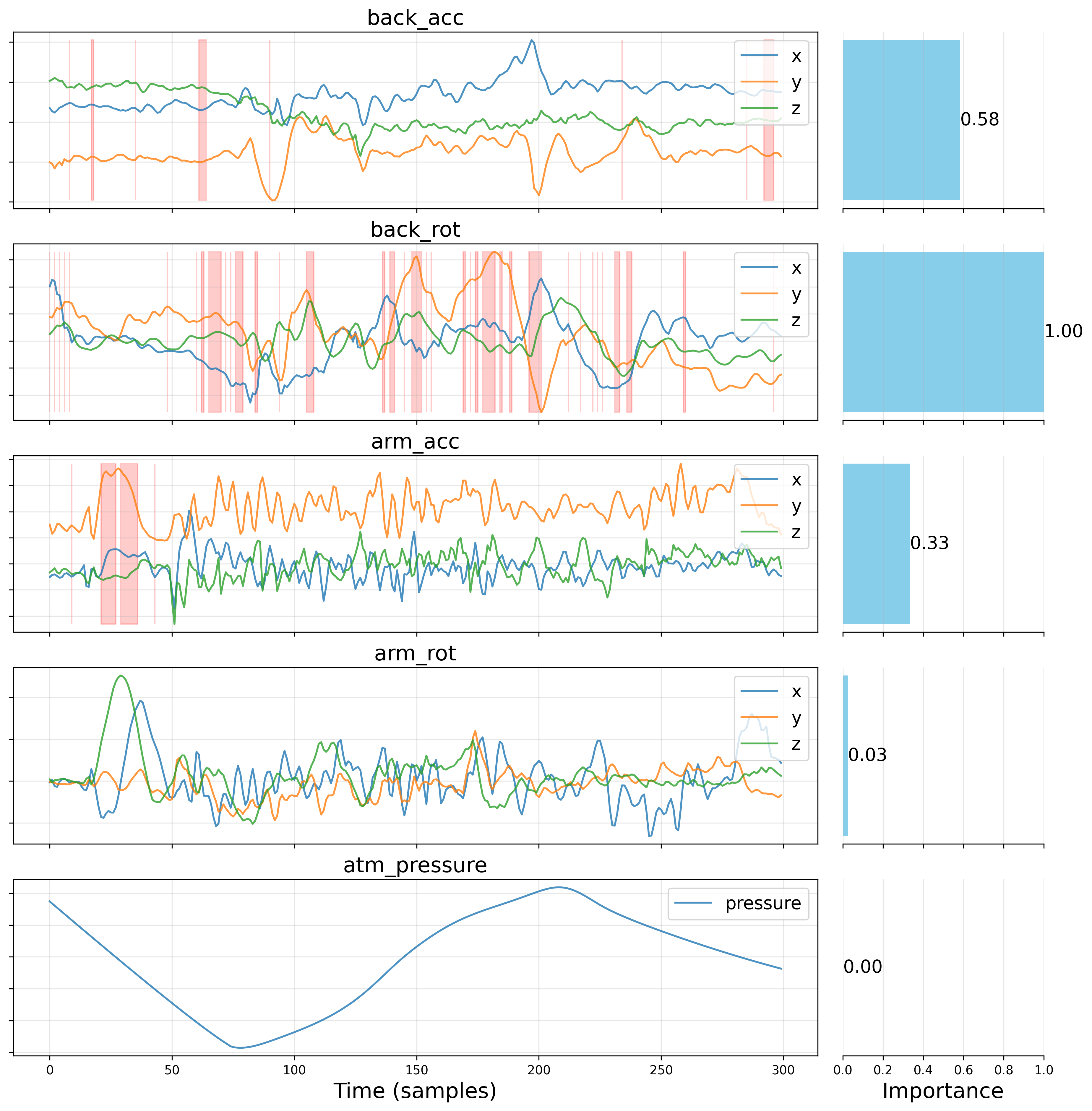} 
    \caption{OpenNet: \textit{cycling}}
    \label{subfig:oppennet_cycling}
  \end{subfigure}

  \vspace{0.5cm} 
  
  \begin{subfigure}[b]{0.45\textwidth}
    \centering
    \includegraphics[width=\textwidth]{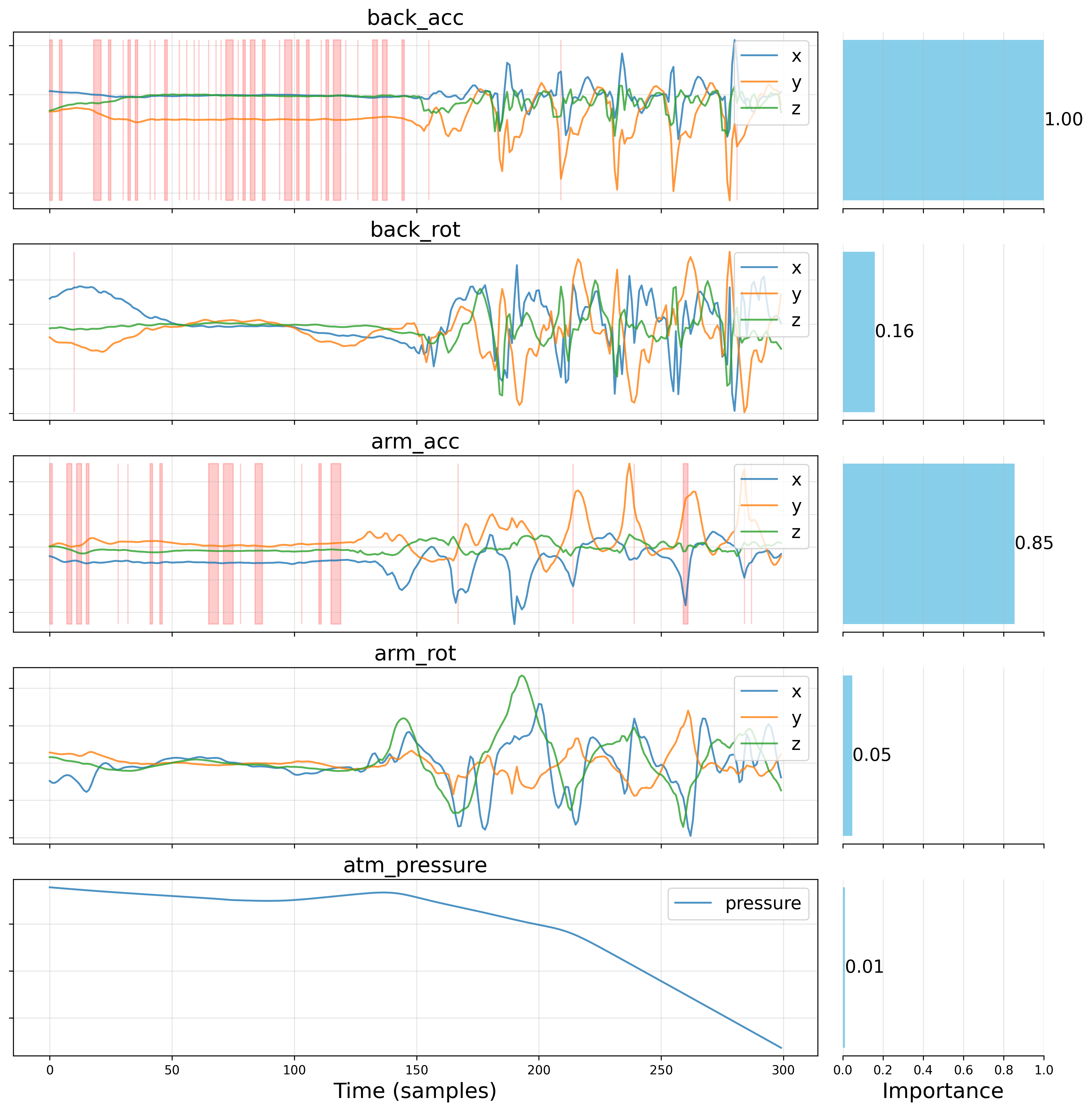} 
    \caption{\modelname: \textit{running}}
    \label{subfig:ours_running}
  \end{subfigure}
  \hfill
  \begin{subfigure}[b]{0.45\textwidth}
    \centering
    \includegraphics[width=\textwidth]{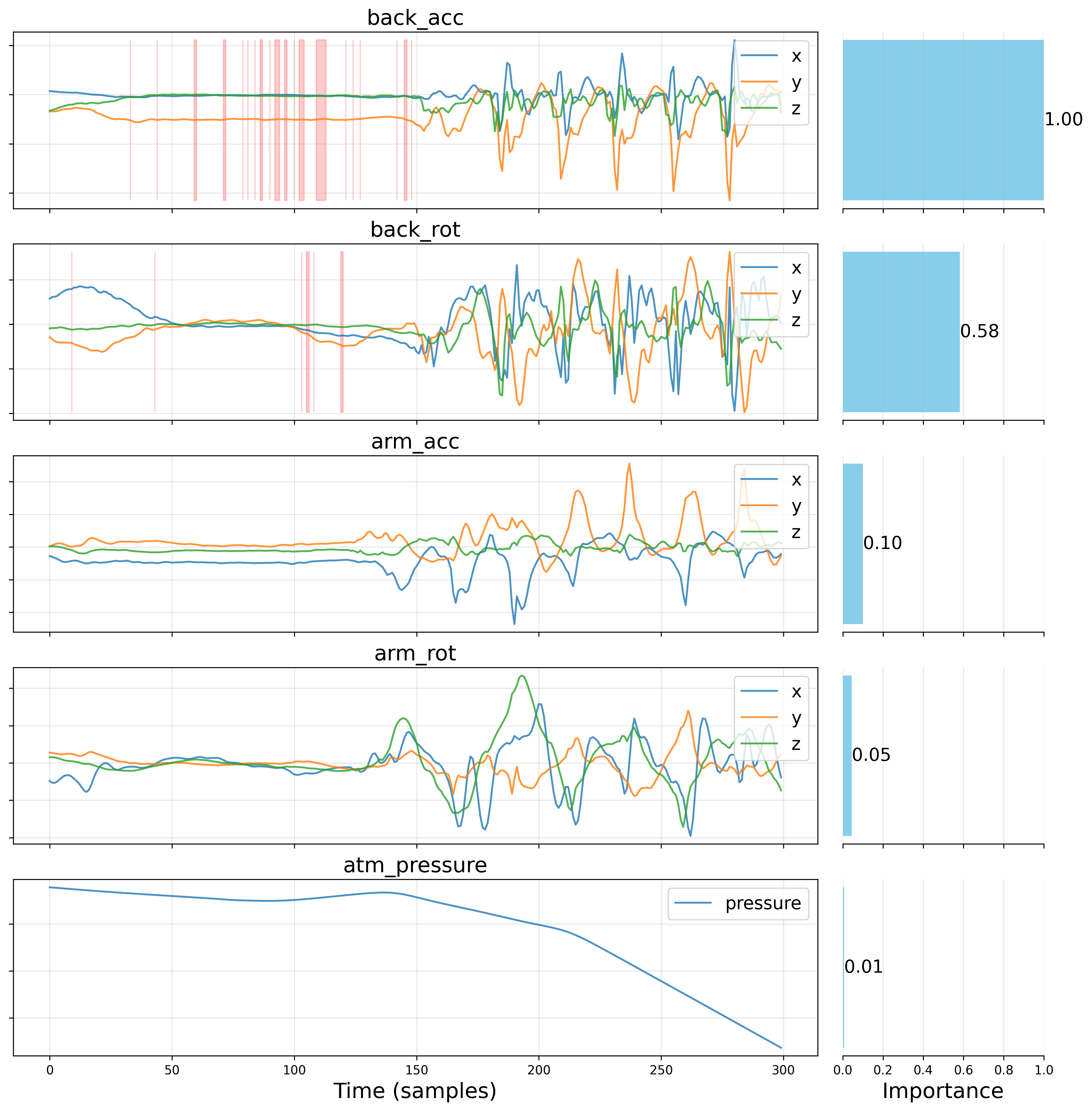} 
    \caption{OpenNet: \textit{running}}
    \label{subfig:opennet_running}
  \end{subfigure}
  
  \caption{Importance of sections of the input sensor data for the output of \modelname vs. OpenNet for classes \textit{cycling} and \textit{running}, calculated using XAI technique Integrated Gradients. Most important sections of the timeseries highlighted in red, and overall sensor importance is shown in the bar graphs to the right. \modelname emphasises the back acceleration and arm acceleration sensors primarily to produce output, while OpenNet emphasises sensors back acceleration and back rotation.}
  \label{fig:sensor_importances}
\end{figure}

To investigate the impact of OOD class on open-set performance, we show the results for \modelname on each class of \datasetname in Table \ref{tab:ood_results_ours}. \datasetname contains a high number of classes compared with existing HAR datasets, and notably, a number of those classes are similar to each other. Overlap or equivalence between classes increases the difficulty of OOD classification for these similar classes. 
\par
OOD results show significant variation. Best performing classes \textit{dragging}, \textit{cycling}, and \textit{standing} all score above 0.85 on AUROC. Meanwhile, worst performing classes \textit{running}, \text{walking}, \textit{punching}, and \textit{stair down} all score below 0.55. Notably, these low scoring classes all have similar activity classes in \datasetname. Across all OOD classes, ID performance is comparatively consistent, with macro-F1 score remaining between 0.812 (\textit{throwing}) and 0.853 (\textit{walking}). 
\par
We compare variations in OOD performance with the best performing OOD baseline, OpenNet in Figure \ref{fig:auroc_comp_hist}. \modelname performs better on 13 out of 20 activities (including NULL from OPPORTUNITY), and is superior on average. Distribution of OOD performance between the methods on many classes is roughly in line for a number of classes, but there are significant differences in some cases. 
\par
In order to investigate the cause of the gaps in OOD performance, we applied the eXplainable AI (XAI) technique Integrated Gradients \cite{sundararajan_axiomatic_2017} to both \modelname and OpenNet. Outputs of Integrated Gradients are shown in Figure \ref{fig:sensor_importances} for classes \textit{cycling} and \textit{running}, the two classes with the largest gap in AUROC in favour of either method. For all cases, the back acceleration sensor is important, and aside from OpenNet: \textit{cycling}, is the most important sensor overall. For \modelname, back acceleration and arm acceleration were the two most important sensors, with less contribution coming from back and arm rotation or atmospheric pressure. For OpenNet, back acceleration and back rotation had the largest contributions, with back rotation being the most important for OOD class \textit{cycling}. Interestingly, this is the OOD class OpenNet performed worst on, relative to \modelname. 
\par
Despite using the same feature extractor, the two methods focus on different parts of input to determine output. Not only are different sections of the input data used more, but across the sensors themselves, \modelname looks more at acceleration from both device placements, while OpenNet uses more acceleration and rotation from the back-placed device. These differences influence the variations in OOD performance seen in Figure \ref{fig:auroc_comp_hist}. 
\subsection{Effect of Threshold on Approximating OOD Class \label{section:experiment:threshold_ward_d}}

\begin{figure}[ht]

    \centering
    \begin{subfigure}[b]{0.45\textwidth}
      \centering
      \includegraphics[width=\textwidth]{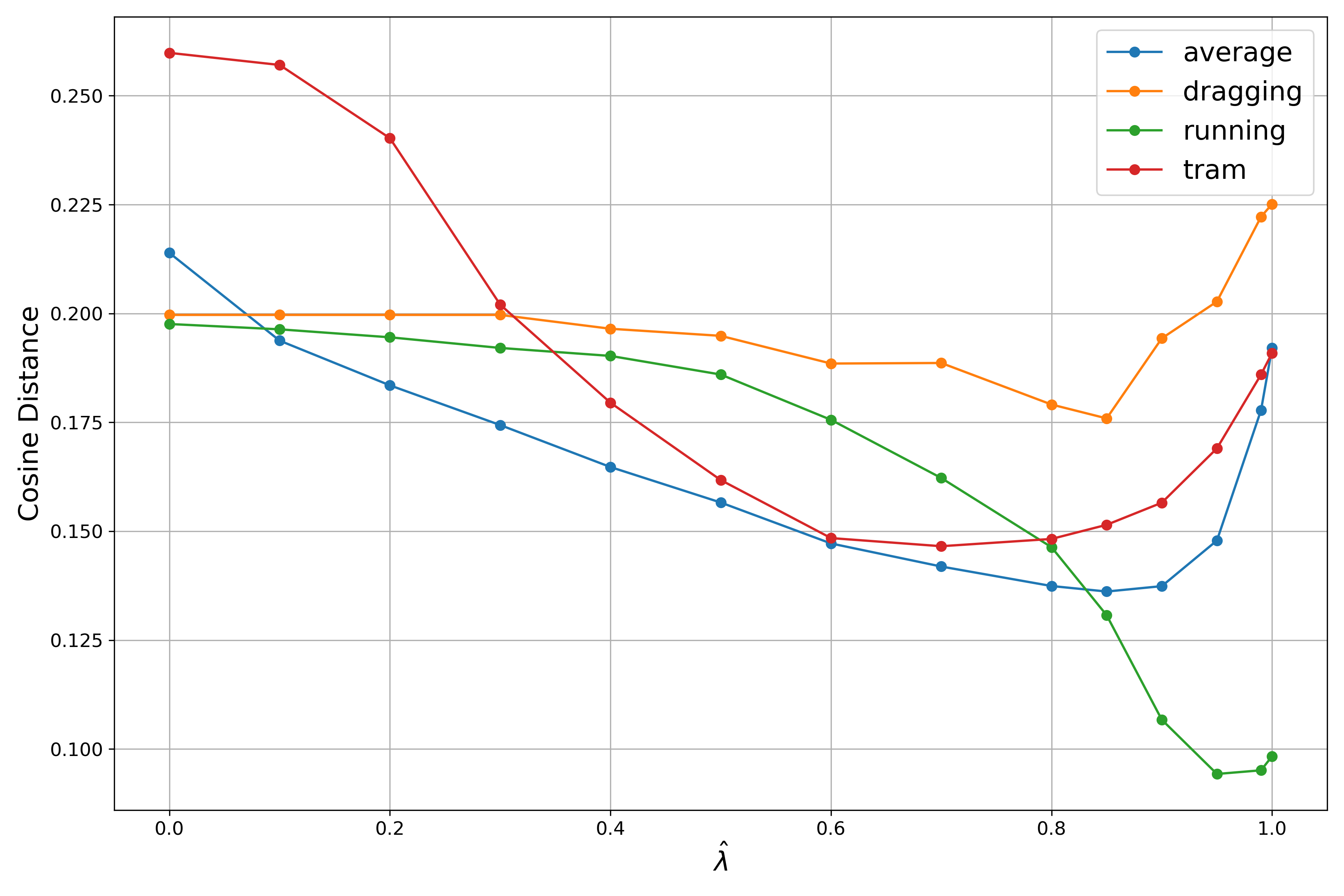} 
      \caption{}
      \label{subfig:cos_d}
    \end{subfigure}
    \hfill
    \begin{subfigure}[b]{0.5\textwidth}
      \centering
      \includegraphics[width=\textwidth]{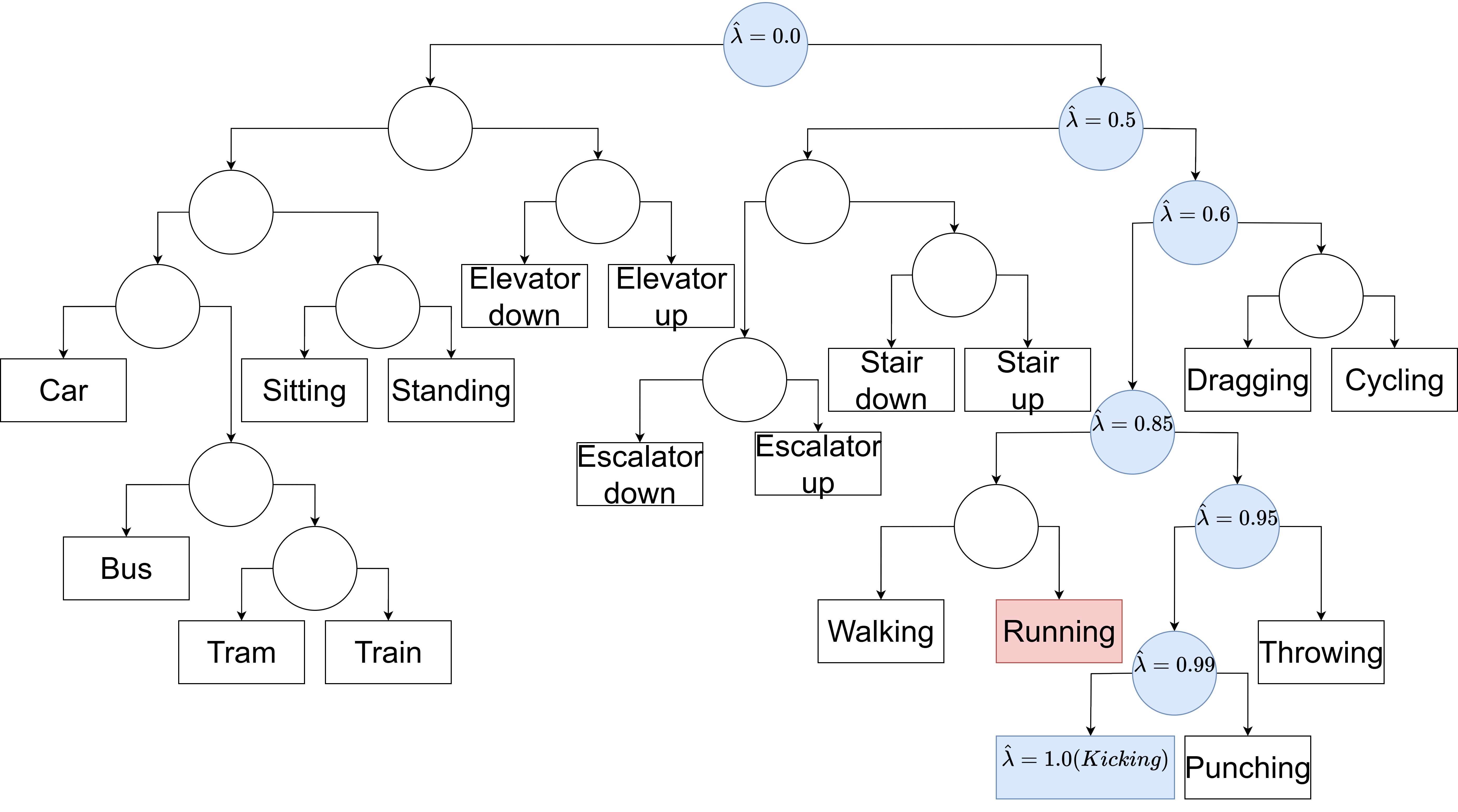} 
      \caption{}
      \label{subfig:running_lambda_preds}
    \end{subfigure}

    \caption{(a) Cumulative Cosine Distance of predicted node from true node position for different OOD classes when varying threshold $\hat{\lambda}$. (b) Example of how predictions (marked in blue) change with $\hat{\lambda}$. The hierarchy was computed using all classes and OOD class \textit{running} was removed during training. Its position in the original hierarchy is marked in red.}
    \label{fig:threshold_cos_d}

\end{figure}    

An added benefit of the hierarchical output is the ability to predict internal nodes for approximating OOD classes. In this experiment, a hierarchy was generated using all classes in \datasetname, so that the position of the predicted internal nodes could be compared to where the OOD class "should" be.  The measure of the quality of a prediction is how close the predicted node is to the true position of the OOD class in the hierarchy. This closeness is measured using Cumulative Cosine Distance (Equation \ref{equation:cum_cos_distance}), where lower distances indicate the predicted node for the OOD is close to its true position. 
\par
Figure \ref{subfig:cos_d} plots threshold $\hat{\lambda}$ against average Cumulative Cosine Distance across all classes, as well as three highlighted classes \textit{dragging}, \textit{running}, and \textit{tram}. High values of $\hat{\lambda}$ return predictions closer to leaf nodes, and as it is reduced, traversal is more likely to terminate earlier up the tree, closer to the root. The blue line in Figure \ref{fig:threshold_cos_d} showing average Cumulative Cosine Distance shows a steady drop from $\hat{\lambda} = 1.00$ to $\hat{\lambda} = 0.85$ before increasing as $\hat{\lambda}$ is reduced further. This corresponds to "overprediction" when $\hat{\lambda}$ is at a maximum, that is, going past the OOD class towards another leaf node, then as $\hat{\lambda}$ is reduced somewhat, the prediction improves, moving up a level or two, closer to the true prediction. As $\hat{\lambda}$ decreases further, predictions continue getting further away from the true position. This is illustrated for OOD class \textit{running} in Figure \ref{subfig:running_lambda_preds}. 
\par
All three plotted classes exhibit an initial decrease before steady increase in distance from true class. The class \textit{tram} follows the shape of the average curve closely, albeit with a flatter portion between $\hat{\lambda} = 0.85$ and $\hat{\lambda} = 0.6$, indicating that predictions do not improve or disimprove much in this range. Class \textit{running}, also shows an improvement in prediction quality as $\hat{\lambda}$ moves from 1.0 to 0.95, before a sharp increase. Notably, \textit{running} was the worst classified OOD class in Section \ref{section:experiments:ood_detection_different}, and here the cosine distance to the true position is very low, revealing that running is being incorrectly classified as a very similar class. \textit{Dragging} is actually closer to the true position when $\hat{\lambda} = 0$ than it is for $\hat{\lambda}=1$. This is an example of when the 'true' class is closer to the Root than nearby leaf nodes, still the minimum distance to the true class occurs at $\hat{\lambda} = 0.85$. Although the curves are show some differences in shape and minimum, setting $\hat{\lambda} = 0.85$, the optimal setting on average, for all curves returns predictions close to their optimal values.
\subsection{Alternative Hierarchies \label{section:experiment:diff_hierarchies}}
\begin{table}
\begin{center}
    
\begin{tabular}{c|c|c|c}
\hline
Hierarchy Type & AUROC $\uparrow$ & Detection Error $\downarrow$ & F1 $\uparrow$  \\
\hline
Chat-GPT 4.5* & 0.624 (0.099) & 0.363 (0.086) & \underline{0.830} (0.037)  \\
Gemini 2.5* & \underline{0.708} (0.080) & 0.337 (0.078) & \underline{0.820} (0.041)  \\
WordNet* & 0.616 (0.068) & 0.378 (0.063) & \underline{0.819} (0.039)  \\
Handcrafted & \underline{0.735} (0.129) & \underline{0.300} (0.100) & \underline{0.831} (0.038)  \\
ECDF & \underline{0.743} (0.117) & \underline{0.298} (0.091) & \textbf{0.832} (0.036)  \\
Feature Embeddings & \textbf{0.774} (0.104) & \textbf{0.263} (0.084) & \underline{0.830} (0.040)  \\
\hline
\end{tabular}
\caption{Average metrics for OOD classes \textit{tram}, \textit{cycling}, and \textit{sitting} from \datasetname for different hierarchy generation methods. Methods marked with an asterix * are knowledge-driven, while the others are generated using data from the training set. Best results are in bold. Underlined scores indicate models whose performance is not statistically significantly different (p $\geq$ 0.05) from the best-performing model, according to a paired t-test over the folds.\label{tab:diff_hierarchies}}
\end{center}
\end{table}


The choice of hierarchy generation method also has an influence on the performance of \modelname. To gauge this influence, we tested a number of different hierarchy generation methods. Three data-driven methods were included, which used HAC as described in Section \ref{section:method:hierarchy_generation} with different input features to generate the hierarchy. The \textit{Feature Embeddings} method uses the features from the embedding space as described in Section \ref{section:method:hierarchy_generation}, \textit{ECDF} uses the Empirical Cumulative Distribution Function for each axis, and \textit{Handcrafted} uses handcrafted features such as mean, variance, skew, kurtosis, etc. from the windows of data. All features are calculated using only training samples. The three knowledge-driven hierarchies are generated purely from the class labels. Methods \textit{Chat-GPT 4.5} and \textit{Gemini 2.5} are generated by feeding the list of training classes into the respective LLMs and asking for a hierarchy to be produced. The prompt used is provided in Appendix \ref{section:appendix:llm_prompt}. The \textit{WordNet} hierarchy is created using the hierarchical relations from the WordNet \cite{fellbaum_wordnet_2010} database. Classes \textit{cycling}, \textit{tram}, and \textit{sitting} from \datasetname were used individually as OOD classes and results averaged and displayed in Table \ref{tab:diff_hierarchies}. The selected classes were chosen for being dissimilar in nature and assessing different aspects of the methods' OOD detection to give an indication of overall performance
\par
The best performing hierarchy in terms of OOD detection was Feature Embeddings, scoring 0.774 and 0.263 for AUROC and detection error, respectively. However, the two other data-driven methods were only marginally worse, and were not significantly different according to a paired t-test. Gemini 2.5 was the best knowledge-driven hierarchy, and was not significantly worse, although the effect size was quite large at -0.8. All hierarchies had similar F1-scores. This indicates that hierarchy choice effects primarily OOD detection in \modelname, and has minimal impact on ID performance.
\subsection{Ablation Study \label{section:experiment:ablation}}
\begin{table}
\begin{center}
    
\begin{tabular}{c c|c|c|c}
\hline
Hierarchy & SSL & AUROC $\uparrow$ & Detection Error $\downarrow$ & F1 $\uparrow$  \\
\hline
\ding{55} & \ding{55} & 0.578 (0.110) & 0.385 (0.041) & 0.658 (0.086) \\
\ding{51} & \ding{55} & 0.554 (0.138) & 0.387 (0.046) & 0.686 (0.079) \\
\ding{55} & \ding{51} & 0.683 (0.087) & 0.310 (0.080) & 0.833 (0.038) \\
\ding{51} & \ding{51} & 0.774 (0.104) & 0.263 (0.084) & 0.830 (0.040) \\
\hline
\end{tabular}
\caption{Average metrics for OOD classes \textit{tram}, \textit{cycling}, and \textit{sitting} from \datasetname using different components of the \modelname architecture. SSL \ding{55} does not use the pretrained weights and initialises them randomly, and Hierarchy \ding{55} replaces the hierarchy component of with a single softmax layer. Loss functions are adjusted accordingly.
\label{tab:ablation}}
\end{center}
\end{table}

To test the contribution of the different components of Hi-OSCAR, experiments were repeated for OOD classes \textit{tram}, \textit{cycling}, and \textit{sitting} ablating over the hierarchy and pretrained weights from Yuan et al. \cite{yuan_self-supervised_2022}. ID performance was consistent between the ID and OOD experiments, so the ID metrics reported here are reflective of the overall ID performance. The hierarchy was replaced with a single layer of $|K|$ nodes. Consequently, only $L_{ID}$ (\ref{eqn:l_id}) could be used as the loss function. To ablate over the self-supervised pretrained weights, the feature extractor is randomly initialised and trained from scratch. Results are displayed in Table \ref{tab:ablation}.
\par
Training the feature extractor from scratch has the biggest impact on both OOD and ID performance, with results far below both setups employing the pretrained weights. Training from scratch also reduces the quality of the generated hierarchy, and is not able to create a sufficiently representative hierarchy. This hampers OOD performance as a consequence. When combined with pretraining, the hierarchy improves OOD performance by a significant margin compared to using no hierarchy at all. However, ID performance between the hierarchical and non-hierarchical models is equivalent. The primary performance benefit of the hierarchy is in OOD detection, as well as the OOD class approximation enabled by the hierarchy. 
\subsection{Impact of Window Size \label{section:experiment:window_size}}
\begin{table}[ht]
\centering
\begin{tabular}{c|c|c|c|c}
\hline 
\multicolumn{2}{c|}{} & \multicolumn{2}{c|}{OOD} & \multicolumn{1}{c}{ID} \\
\hline
Window length & Dataset & AUROC $\uparrow$ & Detection Error $\downarrow$ & F1 $\uparrow$  \\

\hline
\multirow{2}{*}{5 seconds} 
  & \datasetname & 0.681 (0.115) & 0.322 (0.063) & 0.727 (0.048) \\
  \cline{2-5} 
  & OPPORTUNITY & 0.523 (0.356) & 0.201 (0.144) & 0.797 (0.044) \\
\hline 
\multirow{2}{*}{10 seconds} 
  & \datasetname & 0.774 (0.263) & 0.263 (0.101) & 0.833 (0.032) \\
  \cline{2-5} 
  & OPPORTUNITY & 0.820 (0.048) & 0.188 (0.042) & 0.828 (0.050) \\
\hline
\end{tabular}
\caption{Results of using different window sizes for OOD classes \textit{cycling}, \textit{tram}, and \textit{sitting} from \datasetname, and the NULL class from OPPORTUNITY.}
\label{tab:window_results}
\end{table}


Above experiments were all performed using a window size of ten seconds, which is towards the upper end of typical window lengths utilised in HAR. Table \ref{tab:window_results} compares performance when using instead shorter windows of five seconds, on both \datasetname and OPPORTUNITY. For both datasets, ten second windows perform better on the ID and OOD metrics. This is likely due to two factors: firstly, the longer windows provide more context and information, leading to better results, and secondly, the five second model is a smaller size compared to the ten second, also influencing results. 
\section{Discussion} \label{section:discussion}

\subsection{Open-set Classification \label{section:discussion:opensetclassifier}}

Effective open-set classification requires both sufficient ID and OOD performance. For ID tasks, \modelname outperforms all baselines with the exception of Random Forest on OPPORTUNITY, the smallest tested dataset. While baselines score within ~2\% of \modelname on the datasets PAMAP, OPPORTUNITY, and RealWorld, on the more complex data within \datasetname, there is a large gap in performance, with \modelname far outperforming baselines. Additionally, \modelname has the smallest performance difference from the smaller datasets to the larger \datasetname. This demonstrates a good ability of the model to generalise to different settings. 
\par
On OOD tasks, \modelname again performs well, beating baselines on average. Additionally, although OpenNet scores equally as well as \modelname on OOD tasks, this came at the expense of its ID performance, which is 11\% lower than \modelname on average, which is of course a major problem for open-set classification. Conversely, kNN matches \modelname on ID tasks, but is worse on OOD detection. Our use of a hierarchy allows effective simulation of OOD classes without compromising ID performance, producing a balanced open-set classifier. 
\par
For all OOD methods, performance varies between classes, but interestingly, the variations depend on the method. The methods each have unique difficulties for some OOD classes, which are no problem for others. We find that the two best OOD methods, \modelname and OpenNet, focus more on different parts of the sensor input to reach conclusions. These different focuses are what give rise to the variability in OOD detection. 
\subsection{Approximating OOD classes}

Use of a hierarchical classifier in our set up provides the additional benefit of assigning more information to samples labelled OOD, and using HAC with Cosine linkage provides a way for these predictions to be evaluated quantitatively. We show that optimal internal predictions could be returned by setting inference stopping criterion $\hat{\lambda} = 0.85$. This on average returned the closest possible node to the true OOD class. 
\par
The use of Cumulative Cosine Distance as a measure of prediction quality also helps evaluate overall OOD performance. For example, although for \modelname OOD class \textit{running} is weakest, the cosine distances of predictions even when $\hat{\lambda} = 1.00$ are extremely low, meaning that despite predictions being wrong for this class, they are still good approximations of the true value. For many applications of HAR, this can be acceptable, particularly given the overlap between running and similar activities.
\par
It was also demonstrated that different hierarchy generation choices can be used without major loss of performance. Although data-driven hierarchies were overall better due to improved modeling of the inherent physical similarities of activities rather than semantic similarities, Gemini 2.5 was shown to produce an effective hierarchy. This may be useful for a practitioner emphasising semantic interpretation of results. However, adopting such a knowledge-driven method removes the cosine distance metric from the hierarchy, limiting the interpretability of returned internal nodes. 
\subsection{Dataset \datasetname \label{section:discussion:new_dataset}}
This work also sees the introduction of \datasetname, a new HAR dataset. Compared to existing datasets it has more diverse activities and a large amount of data in terms of length. All tested models perform worst on \datasetname ID tasks by a significant margin. The dataset presents an increased challenge that demands more rounded approaches. The similarity between classes requires more fine grain representations to be learned, requiring more robust and adaptable HAR classifiers such as Hi-OSCAR. Additional challenge comes from the data collection procedure. There is deliberate within-class variations such as running and walking up stairs or sitting with and without a backrest. The result is that even individual classes are more difficult to classify by broadening the class definitions to include realistic intra-class diversity.
\par
\datasetname is also utilised for OOD classification, a task with limited feasibility using existing datasets. On this aspect there is still room for improvement. The challenge of the classes themselves, combined with the difficulty in distinguishing OOD classes from overlapping ID classes increases the intricacy of OOD detection. 
\par
The difficulty in OOD detection is apparent in the variability of OOD performance present in all tested methods. Baselines struggle in some cases where there is significant overlap between ID and OOD classes. In general machine learning, this phenomenon is known as near-OOD detection, and is typically much more challenging than far-OOD detection, where OOD classes contain obvious domain shifts from ID classes. Some candidates for near-OOD classes could be determined from the data collection procedure. For example, the similarity between \textit{stair up/down} and \textit{escalator up/down}, is obvious, but also classes  \textit{kicking} and \textit{punching} could be classed near-OOD. The two activities were collected consecutively during one session with a punching bag. Dividing classes into near- and far-OOD could be a promising avenue of research, but also presents a new set of problems to solve. While existing datasets such as OPPORTUNITY can be used for OOD detection with the NULL class, the lack of information about the NULL samples means possible analysis is limited when compared to \datasetname. Overall, there is a considerable deal of further research enabled by \datasetname.
\section{Conclusions and Future Work\label{section:conclusion}} 

In this work we present our new model \modelname for open-set classification on HAR data. \modelname provides the best balance between ID and OOD classification, mimicking the real-world requirements of HAR. This is incorporated into the model architecture and training to improve results, and also provides a continuous measure of similarity between classes which is used to evaluate the quality of predictions. To the best of our knowledge, we are also the first to model HAR classes as a hierarchy and utilise Cosine Distance in this way. Additionally, we introduce \datasetname, a challenging new HAR dataset. In particular, the variation in similarities between activity classes in the dataset increases the difficulty of OOD detection. In future work, we will look to leverage the internal node information to form novel classes and update the hierarchy, forming an adaptive network.

\bibliography{references}

\begin{thebibliography}{10}

\bibitem{al-qaness_multi-resatt_2023}
M.~A.~A. Al-qaness, A.~Dahou, M.~A. Elaziz, and A.~M. Helmi.
\newblock Multi-{ResAtt}: {Multilevel} {Residual} {Network} {With} {Attention} for {Human} {Activity} {Recognition} {Using} {Wearable} {Sensors}.
\newblock {\em IEEE Transactions on Industrial Informatics}, 19(1):144--152, Jan. 2023.

\bibitem{alemayoh_new_2021}
T.~T. Alemayoh, J.~H. Lee, and S.~Okamoto.
\newblock New {Sensor} {Data} {Structuring} for {Deeper} {Feature} {Extraction} in {Human} {Activity} {Recognition}.
\newblock {\em Sensors}, 21(8):2814, Jan. 2021.
\newblock Number: 8 Publisher: Multidisciplinary Digital Publishing Institute.

\bibitem{anguita_public_2013}
D.~Anguita, A.~Ghio, L.~Oneto, X.~Parra, and J.~L. Reyes-Ortiz.
\newblock A {Public} {Domain} {Dataset} for {Human} {Activity} {Recognition} {Using} {Smartphones}.
\newblock {\em Computational Intelligence}, 2013.

\bibitem{arrotta_dexar_2022}
L.~Arrotta, G.~Civitarese, and C.~Bettini.
\newblock {DeXAR}: {Deep} {Explainable} {Sensor}-{Based} {Activity} {Recognition} in {Smart}-{Home} {Environments}.
\newblock {\em Proceedings of the ACM on Interactive, Mobile, Wearable and Ubiquitous Technologies}, 6(1):1:1--1:30, Mar. 2022.

\bibitem{arrotta_semantic_2024}
L.~Arrotta, G.~Civitarese, and C.~Bettini.
\newblock Semantic {Loss}: {A} {New} {Neuro}-{Symbolic} {Approach} for {Context}-{Aware} {Human} {Activity} {Recognition}.
\newblock {\em Proceedings of the ACM on Interactive, Mobile, Wearable and Ubiquitous Technologies}, 7(4):147:1--147:29, Jan. 2024.

\bibitem{asim_context-aware_2020}
Y.~Asim, M.~A. Azam, M.~Ehatisham-ul Haq, U.~Naeem, and A.~Khalid.
\newblock Context-{Aware} {Human} {Activity} {Recognition} ({CAHAR}) in-the-{Wild} {Using} {Smartphone} {Accelerometer}.
\newblock {\em IEEE Sensors Journal}, 20(8):4361--4371, Apr. 2020.
\newblock Conference Name: IEEE Sensors Journal.

\bibitem{attal_physical_2015}
F.~Attal, S.~Mohammed, M.~Dedabrishvili, F.~Chamroukhi, L.~Oukhellou, and Y.~Amirat.
\newblock Physical {Human} {Activity} {Recognition} {Using} {Wearable} {Sensors}.
\newblock {\em Sensors}, 15(12):31314--31338, Dec. 2015.
\newblock Number: 12 Publisher: Multidisciplinary Digital Publishing Institute.

\bibitem{balli_human_2019}
S.~Balli, E.~A. Sağbaş, and M.~Peker.
\newblock Human activity recognition from smart watch sensor data using a hybrid of principal component analysis and random forest algorithm.
\newblock {\em Measurement and Control}, 52(1-2):37--45, Jan. 2019.
\newblock Publisher: SAGE Publications Ltd.

\bibitem{bartolome_glucomine_2021}
A.~Bartolome, S.~Shah, and T.~Prioleau.
\newblock {GlucoMine}: {A} {Case} for {Improving} the {Use} of {Wearable} {Device} {Data} in {Diabetes} {Management}.
\newblock {\em Proceedings of the ACM on Interactive, Mobile, Wearable and Ubiquitous Technologies}, 5(3):90:1--90:24, Sept. 2021.

\bibitem{bayat_study_2014}
A.~Bayat, M.~Pomplun, and D.~A. Tran.
\newblock A {Study} on {Human} {Activity} {Recognition} {Using} {Accelerometer} {Data} from {Smartphones}.
\newblock {\em Procedia Computer Science}, 34:450--457, 2014.

\bibitem{banos_benchmark_2012}
O.~Baños, M.~Damas, H.~Pomares, I.~Rojas, M.~A. Tóth, and O.~Amft.
\newblock A benchmark dataset to evaluate sensor displacement in activity recognition.
\newblock In {\em Proceedings of the 2012 {ACM} {Conference} on {Ubiquitous} {Computing} - {UbiComp} '12}, page 1026, Pittsburgh, Pennsylvania, 2012. ACM Press.

\bibitem{belcher_us_2021}
B.~R. Belcher, D.~L. Wolff-Hughes, E.~E. Dooley, J.~Staudenmayer, D.~Berrigan, M.~S. Eberhardt, and R.~P. Troiano.
\newblock U.{S}. {Population}-referenced {Percentiles} for {Wrist}-{Worn} {Accelerometer}-derived {Activity}.
\newblock {\em Medicine and science in sports and exercise}, 53(11):2455--2464, Nov. 2021.

\bibitem{bendale_towards_2016}
A.~Bendale and T.~E. Boult.
\newblock Towards {Open} {Set} {Deep} {Networks}.
\newblock pages 1563--1572, 2016.

\bibitem{berchtold_actiserv_2010}
M.~Berchtold, M.~Budde, D.~Gordon, H.~R. Schmidtke, and M.~Beigl.
\newblock {ActiServ}: {Activity} {Recognition} {Service} for mobile phones.
\newblock In {\em International {Symposium} on {Wearable} {Computers} ({ISWC}) 2010}, pages 1--8, Oct. 2010.
\newblock ISSN: 2376-8541.

\bibitem{boyer_out--distribution_2021}
P.~Boyer, D.~Burns, and C.~Whyne.
\newblock Out-of-{Distribution} {Detection} of {Human} {Activity} {Recognition} with {Smartwatch} {Inertial} {Sensors}.
\newblock {\em Sensors}, 21(5):1669, Jan. 2021.
\newblock Number: 5 Publisher: Multidisciplinary Digital Publishing Institute.

\bibitem{chatzaki_human_2017}
C.~Chatzaki, M.~Pediaditis, G.~Vavoulas, and M.~Tsiknakis.
\newblock Human {Daily} {Activity} and {Fall} {Recognition} {Using} a {Smartphone}’s {Acceleration} {Sensor}.
\newblock In C.~Röcker, J.~O'Donoghue, M.~Ziefle, M.~Helfert, and W.~Molloy, editors, {\em Information and {Communication} {Technologies} for {Ageing} {Well} and e-{Health}}, pages 100--118, Cham, 2017. Springer International Publishing.

\bibitem{chavarriaga_opportunity_2013}
R.~Chavarriaga, H.~Sagha, A.~Calatroni, S.~T. Digumarti, G.~Tröster, J.~d.~R. Millán, and D.~Roggen.
\newblock The {Opportunity} challenge: {A} benchmark database for on-body sensor-based activity recognition.
\newblock {\em Pattern Recognition Letters}, 34(15):2033--2042, Nov. 2013.

\bibitem{chen_deep_2021}
L.~Chen, X.~Liu, L.~Peng, and M.~Wu.
\newblock Deep learning based multimodal complex human activity recognition using wearable devices.
\newblock {\em Applied Intelligence}, 51(6):4029--4042, June 2021.

\bibitem{chen_apneadetector_2021}
X.~Chen, Y.~Xiao, Y.~Tang, J.~Fernandez-Mendoza, and G.~Cao.
\newblock {ApneaDetector}: {Detecting} {Sleep} {Apnea} with {Smartwatches}.
\newblock {\em Proceedings of the ACM on Interactive, Mobile, Wearable and Ubiquitous Technologies}, 5(2):59:1--59:22, June 2021.

\bibitem{chen_open-set_2024}
Y.~Chen, W.~Cui, Y.~Huang, C.~Liu, and T.~Zhu.
\newblock Open-{Set} {Sensor} {Human} {Activity} {Recognition} {Based} on {Reciprocal} {Time} {Series}.
\newblock In Z.~Shi, J.~Torresen, and S.~Yang, editors, {\em Intelligent {Information} {Processing} {XII}}, pages 101--115, Cham, 2024. Springer Nature Switzerland.

\bibitem{chen_deep_2015}
Y.~Chen and Y.~Xue.
\newblock A {Deep} {Learning} {Approach} to {Human} {Activity} {Recognition} {Based} on {Single} {Accelerometer}.
\newblock In {\em 2015 {IEEE} {International} {Conference} on {Systems}, {Man}, and {Cybernetics}}, pages 1488--1492, Oct. 2015.

\bibitem{cherian_exploring_2024}
J.~Cherian, S.~Ray, P.~Taele, J.~I. Koh, and T.~Hammond.
\newblock Exploring the {Impact} of the {NULL} {Class} on {In}-the-{Wild} {Human} {Activity} {Recognition}.
\newblock {\em Sensors}, 24(12):3898, Jan. 2024.
\newblock Number: 12 Publisher: Multidisciplinary Digital Publishing Institute.

\bibitem{dirgova_luptakova_wearable_2022}
I.~Dirgová~Luptáková, M.~Kubovčík, and J.~Pospíchal.
\newblock Wearable {Sensor}-{Based} {Human} {Activity} {Recognition} with {Transformer} {Model}.
\newblock {\em Sensors}, 22(5):1911, Jan. 2022.
\newblock Number: 5 Publisher: Multidisciplinary Digital Publishing Institute.

\bibitem{djurisic_extremely_2023}
A.~Djurisic, N.~Bozanic, A.~Ashok, and R.~Liu.
\newblock Extremely {Simple} {Activation} {Shaping} for {Out}-of-{Distribution} {Detection}, May 2023.
\newblock arXiv:2209.09858 [cs].

\bibitem{doherty_large_2017}
A.~Doherty, D.~Jackson, N.~Hammerla, T.~Plötz, P.~Olivier, M.~H. Granat, T.~White, V.~T.~v. Hees, M.~I. Trenell, C.~G. Owen, S.~J. Preece, R.~Gillions, S.~Sheard, T.~Peakman, S.~Brage, and N.~J. Wareham.
\newblock Large {Scale} {Population} {Assessment} of {Physical} {Activity} {Using} {Wrist} {Worn} {Accelerometers}: {The} {UK} {Biobank} {Study}.
\newblock {\em PLOS ONE}, 12(2):e0169649, Feb. 2017.
\newblock Publisher: Public Library of Science.

\bibitem{du_vos_2022}
X.~Du, Z.~Wang, M.~Cai, and Y.~Li.
\newblock {VOS}: {Learning} {What} {You} {Don}'t {Know} by {Virtual} {Outlier} {Synthesis}, May 2022.
\newblock arXiv:2202.01197 [cs].

\bibitem{fellbaum_wordnet_2010}
C.~Fellbaum.
\newblock {WordNet}.
\newblock In R.~Poli, M.~Healy, and A.~Kameas, editors, {\em Theory and {Applications} of {Ontology}: {Computer} {Applications}}, pages 231--243. Springer Netherlands, Dordrecht, 2010.

\bibitem{gao_individual_2022}
N.~Gao, M.~S. Rahaman, W.~Shao, K.~Ji, and F.~D. Salim.
\newblock Individual and {Group}-wise {Classroom} {Seating} {Experience}: {Effects} on {Student} {Engagement} in {Different} {Courses}.
\newblock {\em Proceedings of the ACM on Interactive, Mobile, Wearable and Ubiquitous Technologies}, 6(3):115:1--115:23, Sept. 2022.

\bibitem{gao_n-gage_2020}
N.~Gao, W.~Shao, M.~S. Rahaman, and F.~D. Salim.
\newblock n-{Gage}: {Predicting} in-class {Emotional}, {Behavioural} and {Cognitive} {Engagement} in the {Wild}.
\newblock {\em Proceedings of the ACM on Interactive, Mobile, Wearable and Ubiquitous Technologies}, 4(3):79:1--79:26, Sept. 2020.

\bibitem{gao_diffguard_2023}
R.~Gao, C.~Zhao, L.~Hong, and Q.~Xu.
\newblock {DIFFGUARD}: {Semantic} {Mismatch}-{Guided} {Out}-of-{Distribution} {Detection} {Using} {Pre}-{Trained} {Diffusion} {Models}.
\newblock pages 1579--1589, 2023.

\bibitem{hassen_learning_2020}
M.~Hassen and P.~K. Chan.
\newblock Learning a {Neural}-network-based {Representation} for {Open} {Set} {Recognition}.
\newblock In {\em Proceedings of the 2020 {SIAM} {International} {Conference} on {Data} {Mining} ({SDM})}, Proceedings, pages 154--162. Society for Industrial and Applied Mathematics, Jan. 2020.

\bibitem{helmi_human_2023}
A.~M. Helmi, M.~A.~A. Al-qaness, A.~Dahou, and M.~Abd~Elaziz.
\newblock Human activity recognition using marine predators algorithm with deep learning.
\newblock {\em Future Generation Computer Systems}, 142:340--350, May 2023.

\bibitem{hendrycks_baseline_2018}
D.~Hendrycks and K.~Gimpel.
\newblock A {Baseline} for {Detecting} {Misclassified} and {Out}-of-{Distribution} {Examples} in {Neural} {Networks}, Oct. 2018.
\newblock arXiv:1610.02136 [cs].

\bibitem{hendrycks_deep_2019}
D.~Hendrycks, M.~Mazeika, and T.~Dietterich.
\newblock Deep {Anomaly} {Detection} with {Outlier} {Exposure}, Jan. 2019.
\newblock arXiv:1812.04606 [cs].

\bibitem{huang_importance_2021}
R.~Huang, A.~Geng, and Y.~Li.
\newblock On the {Importance} of {Gradients} for {Detecting} {Distributional} {Shifts} in the {Wild}.
\newblock In {\em Advances in {Neural} {Information} {Processing} {Systems}}, volume~34, pages 677--689. Curran Associates, Inc., 2021.

\bibitem{jain_multi-class_2014}
L.~P. Jain, W.~J. Scheirer, and T.~E. Boult.
\newblock Multi-class {Open} {Set} {Recognition} {Using} {Probability} of {Inclusion}.
\newblock In D.~Fleet, T.~Pajdla, B.~Schiele, and T.~Tuytelaars, editors, {\em Computer {Vision} – {ECCV} 2014}, pages 393--409, Cham, 2014. Springer International Publishing.

\bibitem{jennings_interpreting_2023}
L.~Jennings, M.~Sorell, and H.~G. Espinosa.
\newblock Interpreting the location data extracted from the {Apple} {Health} database.
\newblock {\em Forensic Science International: Digital Investigation}, 44:301504, Mar. 2023.

\bibitem{ji_hargpt_2024}
S.~Ji, X.~Zheng, and C.~Wu.
\newblock {HARGPT}: {Are} {LLMs} {Zero}-{Shot} {Human} {Activity} {Recognizers}?, Mar. 2024.
\newblock arXiv:2403.02727 [cs].

\bibitem{jia_swingnet_2021}
H.~Jia, J.~Hu, and W.~Hu.
\newblock {SwingNet}: {Ubiquitous} {Fine}-{Grained} {Swing} {Tracking} {Framework} via {Stochastic} {Neural} {Architecture} {Search} and {Adversarial} {Learning}.
\newblock {\em Proceedings of the ACM on Interactive, Mobile, Wearable and Ubiquitous Technologies}, 5(3):106:1--106:21, Sept. 2021.

\bibitem{jung_lax-score_2021}
W.~Jung, A.~Watson, S.~Kuehn, E.~Korem, K.~Koltermann, M.~Sun, S.~Wang, Z.~Liu, and G.~Zhou.
\newblock {LAX}-{Score}: {Quantifying} {Team} {Performance} in {Lacrosse} and {Exploring} {IMU} {Features} towards {Performance} {Enhancement}.
\newblock {\em Proceedings of the ACM on Interactive, Mobile, Wearable and Ubiquitous Technologies}, 5(3):109:1--109:28, Sept. 2021.

\bibitem{kay_kinetics_2017}
W.~Kay, J.~Carreira, K.~Simonyan, B.~Zhang, C.~Hillier, S.~Vijayanarasimhan, F.~Viola, T.~Green, T.~Back, P.~Natsev, M.~Suleyman, and A.~Zisserman.
\newblock The {Kinetics} {Human} {Action} {Video} {Dataset}, May 2017.
\newblock arXiv:1705.06950 [cs].

\bibitem{kilgarriff_review_2000}
A.~Kilgarriff.
\newblock Review of {WordNet}: {An} {Electronic} {Lexical} {Database}.
\newblock {\em Language}, 76(3):706--708, 2000.
\newblock Publisher: Linguistic Society of America.

\bibitem{kose_online_nodate}
M.~Kose, O.~D. Incel, and C.~Ersoy.
\newblock Online {Human} {Activity} {Recognition} on {Smart} {Phones}.

\bibitem{kuehne_hmdb_2011}
H.~Kuehne, H.~Jhuang, E.~Garrote, T.~Poggio, and T.~Serre.
\newblock {HMDB}: {A} large video database for human motion recognition.
\newblock In {\em 2011 {International} {Conference} on {Computer} {Vision}}, pages 2556--2563, Nov. 2011.
\newblock ISSN: 2380-7504.

\bibitem{kwapisz_activity_2011}
J.~R. Kwapisz, G.~M. Weiss, and S.~A. Moore.
\newblock Activity recognition using cell phone accelerometers.
\newblock {\em ACM SIGKDD Explorations Newsletter}, 12(2):74--82, Mar. 2011.

\bibitem{li_multiresolution_2023}
J.~Li, H.~Xu, and Y.~Wang.
\newblock Multiresolution {Fusion} {Convolutional} {Network} for {Open} {Set} {Human} {Activity} {Recognition}.
\newblock {\em IEEE Internet of Things Journal}, 10(13):11369--11382, July 2023.
\newblock Conference Name: IEEE Internet of Things Journal.

\bibitem{liang_enhancing_2020}
S.~Liang, Y.~Li, and R.~Srikant.
\newblock Enhancing {The} {Reliability} of {Out}-of-distribution {Image} {Detection} in {Neural} {Networks}, Aug. 2020.
\newblock arXiv:1706.02690 [cs].

\bibitem{linderman_fine-grain_2022}
R.~Linderman, J.~Zhang, N.~Inkawhich, H.~Li, and Y.~Chen.
\newblock Fine-grain {Inference} on {Out}-of-{Distribution} {Data} with {Hierarchical} {Classification}, Sept. 2022.
\newblock arXiv:2209.04493 [cs].

\bibitem{liu_energy-based_2020}
W.~Liu, X.~Wang, J.~Owens, and Y.~Li.
\newblock Energy-based {Out}-of-distribution {Detection}.
\newblock In {\em Advances in {Neural} {Information} {Processing} {Systems}}, volume~33, pages 21464--21475. Curran Associates, Inc., 2020.

\bibitem{malekzadeh_protecting_2018}
M.~Malekzadeh, R.~G. Clegg, A.~Cavallaro, and H.~Haddadi.
\newblock Protecting {Sensory} {Data} against {Sensitive} {Inferences}.
\newblock In {\em Proceedings of the 1st {Workshop} on {Privacy} by {Design} in {Distributed} {Systems}}, W-{P2DS}'18, pages 1--6, New York, NY, USA, Apr. 2018. Association for Computing Machinery.

\bibitem{mekruksavanich_deep_2021}
S.~Mekruksavanich and A.~Jitpattanakul.
\newblock Deep {Convolutional} {Neural} {Network} with {RNNs} for {Complex} {Activity} {Recognition} {Using} {Wrist}-{Worn} {Wearable} {Sensor} {Data}.
\newblock {\em Electronics}, 10(14):1685, Jan. 2021.
\newblock Number: 14 Publisher: Multidisciplinary Digital Publishing Institute.

\bibitem{mekruksavanich_resnet-se_2022}
S.~Mekruksavanich, A.~Jitpattanakul, K.~Sitthithakerngkiet, P.~Youplao, and P.~Yupapin.
\newblock {ResNet}-{SE}: {Channel} {Attention}-{Based} {Deep} {Residual} {Network} for {Complex} {Activity} {Recognition} {Using} {Wrist}-{Worn} {Wearable} {Sensors}.
\newblock {\em IEEE Access}, 10:51142--51154, 2022.
\newblock Conference Name: IEEE Access.

\bibitem{micucci_unimib_2017}
D.~Micucci, M.~Mobilio, and P.~Napoletano.
\newblock {UniMiB} {SHAR}: {A} {Dataset} for {Human} {Activity} {Recognition} {Using} {Acceleration} {Data} from {Smartphones}.
\newblock {\em Applied Sciences}, 7(10):1101, Oct. 2017.
\newblock Number: 10 Publisher: Multidisciplinary Digital Publishing Institute.

\bibitem{mohsen_human_2022}
S.~Mohsen, A.~Elkaseer, and S.~G. Scholz.
\newblock Human {Activity} {Recognition} {Using} {K}-{Nearest} {Neighbor} {Machine} {Learning} {Algorithm}.
\newblock In S.~G. Scholz, R.~J. Howlett, and R.~Setchi, editors, {\em Sustainable {Design} and {Manufacturing}}, pages 304--313, Singapore, 2022. Springer.

\bibitem{murahari_attention_2018}
V.~S. Murahari and T.~Plötz.
\newblock On attention models for human activity recognition.
\newblock In {\em Proceedings of the 2018 {ACM} {International} {Symposium} on {Wearable} {Computers}}, pages 100--103, Singapore Singapore, Oct. 2018. ACM.

\bibitem{nalisnick_deep_2019}
E.~Nalisnick, A.~Matsukawa, Y.~W. Teh, D.~Gorur, and B.~Lakshminarayanan.
\newblock Do {Deep} {Generative} {Models} {Know} {What} {They} {Don}'t {Know}?, Feb. 2019.
\newblock arXiv:1810.09136 [stat].

\bibitem{nie_out--distribution_2025}
J.~Nie, Y.~Luo, S.~Ye, Y.~Zhang, X.~Tian, and Z.~Fang.
\newblock Out-of-{Distribution} {Detection} with {Virtual} {Outlier} {Smoothing}.
\newblock {\em International Journal of Computer Vision}, 133(2):724--741, Feb. 2025.

\bibitem{nurwulan_human_2021}
N.~R. Nurwulan and G.~Selamaj.
\newblock Human daily activities recognition using decision tree.
\newblock {\em Journal of Physics: Conference Series}, 1833(1):012039, Mar. 2021.
\newblock Publisher: IOP Publishing.

\bibitem{okita_towards_2023}
T.~Okita, K.~Ukita, K.~Matsuishi, M.~Kagiyama, K.~Hirata, and A.~Miyazaki.
\newblock Towards {LLMs} for {Sensor} {Data}: {Multi}-{Task} {Self}-{Supervised} {Learning}.
\newblock In {\em Adjunct {Proceedings} of the 2023 {ACM} {International} {Joint} {Conference} on {Pervasive} and {Ubiquitous} {Computing} \& the 2023 {ACM} {International} {Symposium} on {Wearable} {Computing}}, pages 499--504, Cancun, Quintana Roo Mexico, Oct. 2023. ACM.

\bibitem{ordonez_deep_2016}
F.~J. Ordóñez and D.~Roggen.
\newblock Deep {Convolutional} and {LSTM} {Recurrent} {Neural} {Networks} for {Multimodal} {Wearable} {Activity} {Recognition}.
\newblock {\em Sensors}, 16(1):115, Jan. 2016.
\newblock Number: 1 Publisher: Multidisciplinary Digital Publishing Institute.

\bibitem{perera_generative-discriminative_2020}
P.~Perera, V.~I. Morariu, R.~Jain, V.~Manjunatha, C.~Wigington, V.~Ordonez, and V.~M. Patel.
\newblock Generative-{Discriminative} {Feature} {Representations} for {Open}-{Set} {Recognition}.
\newblock pages 11814--11823, 2020.

\bibitem{randell_context_2000}
C.~Randell and H.~Muller.
\newblock Context awareness by analysing accelerometer data.
\newblock In {\em Digest of {Papers}. {Fourth} {International} {Symposium} on {Wearable} {Computers}}, pages 175--176, Oct. 2000.

\bibitem{reiss_introducing_2012}
A.~Reiss and D.~Stricker.
\newblock Introducing a {New} {Benchmarked} {Dataset} for {Activity} {Monitoring}.
\newblock In {\em 2012 16th {International} {Symposium} on {Wearable} {Computers}}, pages 108--109, June 2012.
\newblock ISSN: 2376-8541.

\bibitem{shoaib_fusion_2014}
M.~Shoaib, S.~Bosch, O.~D. Incel, H.~Scholten, and P.~J.~M. Havinga.
\newblock Fusion of {Smartphone} {Motion} {Sensors} for {Physical} {Activity} {Recognition}.
\newblock {\em Sensors}, 14(6):10146--10176, June 2014.
\newblock Number: 6 Publisher: Multidisciplinary Digital Publishing Institute.

\bibitem{shoaib_complex_2016}
M.~Shoaib, S.~Bosch, O.~D. Incel, H.~Scholten, and P.~J.~M. Havinga.
\newblock Complex {Human} {Activity} {Recognition} {Using} {Smartphone} and {Wrist}-{Worn} {Motion} {Sensors}.
\newblock {\em Sensors}, 16(4):426, Apr. 2016.
\newblock Number: 4 Publisher: Multidisciplinary Digital Publishing Institute.

\bibitem{sigurdsson_charades-ego_2018}
G.~A. Sigurdsson, A.~Gupta, C.~Schmid, A.~Farhadi, and K.~Alahari.
\newblock Charades-{Ego}: {A} {Large}-{Scale} {Dataset} of {Paired} {Third} and {First} {Person} {Videos}, Apr. 2018.
\newblock arXiv:1804.09626 [cs].

\bibitem{singh_deep_2021}
S.~P. Singh, M.~K. Sharma, A.~Lay-Ekuakille, D.~Gangwar, and S.~Gupta.
\newblock Deep {ConvLSTM} {With} {Self}-{Attention} for {Human} {Activity} {Decoding} {Using} {Wearable} {Sensors}.
\newblock {\em IEEE Sensors Journal}, 21(6):8575--8582, Mar. 2021.

\bibitem{stisen_smart_2015}
A.~Stisen, H.~Blunck, S.~Bhattacharya, T.~S. Prentow, M.~B. Kjærgaard, A.~Dey, T.~Sonne, and M.~M. Jensen.
\newblock Smart {Devices} are {Different}: {Assessing} and {MitigatingMobile} {Sensing} {Heterogeneities} for {Activity} {Recognition}.
\newblock In {\em Proceedings of the 13th {ACM} {Conference} on {Embedded} {Networked} {Sensor} {Systems}}, pages 127--140, Seoul South Korea, Nov. 2015. ACM.

\bibitem{sundararajan_axiomatic_2017}
M.~Sundararajan, A.~Taly, and Q.~Yan.
\newblock Axiomatic {Attribution} for {Deep} {Networks}.
\newblock In {\em Proceedings of the 34th {International} {Conference} on {Machine} {Learning}}, pages 3319--3328. PMLR, July 2017.
\newblock ISSN: 2640-3498.

\bibitem{sztyler_-body_2016}
T.~Sztyler and H.~Stuckenschmidt.
\newblock On-body localization of wearable devices: {An} investigation of position-aware activity recognition.
\newblock In {\em 2016 {IEEE} {International} {Conference} on {Pervasive} {Computing} and {Communications} ({PerCom})}, pages 1--9, Mar. 2016.

\bibitem{tang_triple_2022}
Y.~Tang, L.~Zhang, Q.~Teng, F.~Min, and A.~Song.
\newblock Triple {Cross}-{Domain} {Attention} on {Human} {Activity} {Recognition} {Using} {Wearable} {Sensors}.
\newblock {\em IEEE Transactions on Emerging Topics in Computational Intelligence}, 6(5):1167--1176, Oct. 2022.
\newblock Conference Name: IEEE Transactions on Emerging Topics in Computational Intelligence.

\bibitem{tao_non-parametric_2023}
L.~Tao, X.~Du, X.~Zhu, and Y.~Li.
\newblock Non-{Parametric} {Outlier} {Synthesis}, Mar. 2023.
\newblock arXiv:2303.02966 [cs].

\bibitem{tayyub_learning_2018}
J.~Tayyub, M.~Hawasly, D.~C. Hogg, and A.~G. Cohn.
\newblock Learning {Hierarchical} {Models} of {Complex} {Daily} {Activities} from {Annotated} {Videos}.
\newblock In {\em 2018 {IEEE} {Winter} {Conference} on {Applications} of {Computer} {Vision} ({WACV})}, pages 1633--1641, Mar. 2018.

\bibitem{tonmoy_hierarchical_2021}
M.~T.~H. Tonmoy, S.~Mahmud, A.~K. M.~M. Rahman, M.~A. Amin, and A.~A. Ali.
\newblock Hierarchical {Self} {Attention} {Based} {Autoencoder} for {Open}-{Set} {Human} {Activity} {Recognition}, Mar. 2021.
\newblock arXiv:2103.04279.

\bibitem{tran_human_2016}
D.~N. Tran and D.~D. Phan.
\newblock Human {Activities} {Recognition} in {Android} {Smartphone} {Using} {Support} {Vector} {Machine}.
\newblock In {\em 2016 7th {International} {Conference} on {Intelligent} {Systems}, {Modelling} and {Simulation} ({ISMS})}, pages 64--68, Bangkok, Thailand, Jan. 2016. IEEE.

\bibitem{van_zandwijk_iphone_2019}
J.~P. van Zandwijk and A.~Boztas.
\newblock The {iPhone} {Health} {App} from a forensic perspective: can steps and distances registered during walking and running be used as digital evidence?
\newblock {\em Digital investigation}, 28:S126--S133, 2019.
\newblock Publisher: Elsevier.

\bibitem{van_zandwijk_phone_2021}
J.~P. van Zandwijk and A.~Boztas.
\newblock The phone reveals your motion: {Digital} traces of walking, driving and other movements on {iPhones}.
\newblock {\em Forensic Science International: Digital Investigation}, 37:301170, June 2021.

\bibitem{wan_nbdt_2021}
A.~Wan, L.~Dunlap, D.~Ho, J.~Yin, S.~Lee, H.~Jin, S.~Petryk, S.~A. Bargal, and J.~E. Gonzalez.
\newblock {NBDT}: {Neural}-{Backed} {Decision} {Trees}, Jan. 2021.
\newblock arXiv:2004.00221.

\bibitem{wang_leveraging_2021}
C.~Wang, Y.~Gao, A.~Mathur, A.~C. De~C.~Williams, N.~D. Lane, and N.~Bianchi-Berthouze.
\newblock Leveraging {Activity} {Recognition} to {Enable} {Protective} {Behavior} {Detection} in {Continuous} {Data}.
\newblock {\em Proceedings of the ACM on Interactive, Mobile, Wearable and Ubiquitous Technologies}, 5(2):81:1--81:27, June 2021.

\bibitem{wang_human_2020}
L.~Wang and R.~Liu.
\newblock Human {Activity} {Recognition} {Based} on {Wearable} {Sensor} {Using} {Hierarchical} {Deep} {LSTM} {Networks}.
\newblock {\em Circuits, Systems, and Signal Processing}, 39(2):837--856, Feb. 2020.

\bibitem{wang_out--distribution_2023}
Q.~Wang, J.~Ye, F.~Liu, Q.~Dai, M.~Kalander, T.~Liu, J.~Hao, and B.~Han.
\newblock Out-of-distribution {Detection} with {Implicit} {Outlier} {Transformation}, Mar. 2023.
\newblock arXiv:2303.05033 [cs].

\bibitem{xu_human_2020}
H.~Xu, J.~Li, H.~Yuan, Q.~Liu, S.~Fan, T.~Li, and X.~Sun.
\newblock Human {Activity} {Recognition} {Based} on {Gramian} {Angular} {Field} and {Deep} {Convolutional} {Neural} {Network}.
\newblock {\em IEEE Access}, 8:199393--199405, 2020.
\newblock Conference Name: IEEE Access.

\bibitem{xu_limu-bert_2021}
H.~Xu, P.~Zhou, R.~Tan, M.~Li, and G.~Shen.
\newblock {LIMU}-{BERT}: {Unleashing} the {Potential} of {Unlabeled} {Data} for {IMU} {Sensing} {Applications}.
\newblock In {\em Proceedings of the 19th {ACM} {Conference} on {Embedded} {Networked} {Sensor} {Systems}}, pages 220--233, Coimbra Portugal, Nov. 2021. ACM.

\bibitem{yao_deepsense_2017}
S.~Yao, S.~Hu, Y.~Zhao, A.~Zhang, and T.~Abdelzaher.
\newblock {DeepSense}: {A} {Unified} {Deep} {Learning} {Framework} for {Time}-{Series} {Mobile} {Sensing} {Data} {Processing}.
\newblock In {\em Proceedings of the 26th {International} {Conference} on {World} {Wide} {Web}}, {WWW} '17, pages 351--360, Republic and Canton of Geneva, CHE, Apr. 2017. International World Wide Web Conferences Steering Committee.

\bibitem{yuan_self-supervised_2022}
H.~Yuan, S.~Chan, A.~P. Creagh, C.~Tong, D.~A. Clifton, and A.~Doherty.
\newblock Self-supervised {Learning} for {Human} {Activity} {Recognition} {Using} 700,000 {Person}-days of {Wearable} {Data}, June 2022.
\newblock arXiv:2206.02909 [cs, eess] version: 1.

\bibitem{zhou_rethinking_2022}
Y.~Zhou.
\newblock Rethinking {Reconstruction} {Autoencoder}-{Based} {Out}-of-{Distribution} {Detection}.
\newblock pages 7379--7387, 2022.

\bibitem{zhu_feature_2017}
J.~Zhu, R.~San-Segundo, and J.~M. Pardo.
\newblock Feature extraction for robust physical activity recognition.
\newblock {\em Human-centric Computing and Information Sciences}, 7(1):16, June 2017.

\end{thebibliography}

\appendix
\section{Appendix}
\subsection{Baseline Hyperparameters \label{section:appendix:hyperparameters}}
Refer to Table \ref{tab:hyperparamters}
\hspace*{-\leftmargin}
\begin{table}[ht]
\small
\scalebox{0.85}{
\begin{tabular}{c|c|c|c}
\hline
Model & Optimiser & Learning Rate & Details  \\
\hline
CNNBiGRU & Adam & 1e-4 & num\_filters: 64, filter\_size: 8, n\_units\_gru: 128, n\_layers\_gru: 2\\ 
Transformer & Adam & 1e-3 &  embedding\_dim: 128, num\_heads: 8, num\_layers: 3, label\_smoothing: 0.1 \\ 
ResNet-SE & Adam & 1e-4 & kernel\_size: 5, conv\_stride: 1, conv\_num\_filters: 64, res\_num\_filters: 32, layer\_size: 128 \\ 
CNNLSTM-Att & RMSProp & 1e-3 & hidden\_size: 128, dropout: 0.6, attention\_dropout: 0.7, num\_layers: 2, num\_filters: 64  \\ 
MLP & Adam & 1e-4 & layer1\_size: 128, layer2\_size: 256, layer3\_size: 512, layer4\_size: 1024  \\ 
Random Forest & n/a & n/a & mean, std, range, med abs dev, kurt, skew, error norm mean. n\_estimators=3000, imblearn default parameters \\ 
\hline

\end{tabular}
}
\caption{Hyperparameters of baselines.}
\label{tab:hyperparamters}
\end{table}
\subsection{LLM Prompt \label{section:appendix:llm_prompt}}

Prompt used to create hierarchies with LLMs Chat-GPT 4.5 and Gemini 2.5: "\textit{[list of activities]}: This is a list of activities that I have in my dataset. Please arrange them in a sensible hierarchy. I want them organised so all activities are part of the same hierarchy i.e. there is a root node that is an ancestor to all activities"
\subsection{Multiple classes in OOD set \label{section:appendix:big_ood_set}}
Refer to Table \ref{tab:big_set_ood_results}
\begin{table}[ht]
\centering
\scalebox{0.85}{
\begin{tabular}{c|c|c|c|c}
\hline 
\multicolumn{2}{c|}{} & \multicolumn{2}{c|}{OOD} & \multicolumn{1}{c}{ID} \\
\hline
Model & OOD set & AUROC $\uparrow$ & Detection Error $\downarrow$ & F1 $\uparrow$  \\
\hline

\multirow{7}{*}{OpenNet \cite{hassen_learning_2020}} 
  & (sitting, bus) & \underline{0.625} (0.132) & \underline{0.387} (0.095) & 0.739 (0.076) \\
  & (stair.down, dragging, throwing) & 0.498 (0.171) & 0.419 (0.071) & 0.662 (0.054) \\
  & (sitting, running, bus) & \textbf{0.668} (0.113) & \textbf{0.360} (0.079) & 0.717 (0.061) \\
  & (cycling, kicking, tram) & \underline{0.730} (0.087) & \underline{0.304} (0.056) & 0.755 (0.096) \\
  & (standing, walking, elevator.up) & 0.632 (0.091) & 0.384 (0.049) & 0.710 (0.060) \\
  & (running, elevator.down, throwing, bus) & \underline{0.628} (0.119) & \underline{0.371} (0.066) & 0.662 (0.096) \\
  & (stair.up, escalator.down, escalator.up, car) & \textbf{0.715} (0.058) & \textbf{0.320} (0.045) & 0.736 (0.059) \\
  
\hline 
\multirow{7}{*}{VAE} 
  & (sitting, bus) & 0.584 (0.098) & \underline{0.396} (0.058) & 0.609 (0.038) \\
  & (stair.down, dragging, throwing) & 0.501 (0.099) & 0.396 (0.068) & 0.575 (0.036) \\
  & (sitting, running, bus) & \underline{0.549} (0.072) & \underline{0.426} (0.041) & 0.533 (0.094) \\
  & (cycling, kicking, tram) & 0.505 (0.098) & 0.427 (0.039) & 0.612 (0.041) \\
  & (standing, walking, elevator.up) & 0.595 (0.063) & 0.402 (0.050) & 0.635 (0.029) \\
  & (running, elevator.down, throwing, bus) & 0.493 (0.133) & 0.424 (0.061) & 0.570 (0.053) \\
  & (stair.up, escalator.down, escalator.up, car) & 0.430 (0.119) & 0.472 (0.032) & 0.634 (0.026) \\

\hline
\multirow{7}{*}{kNN} 
  & (sitting, bus) & 0.355 (0.093) & 0.472 (0.059) & \underline{0.837} (0.060) \\
  & (stair.down, dragging, throwing) & \textbf{0.844} (0.048) & \textbf{0.217} (0.047) & 0.809 (0.051) \\
  & (sitting, running, bus) & 0.510 (0.134) & \underline{0.404} (0.059) & \textbf{0.837} (0.036) \\
  & (cycling, kicking, tram) & 0.571 (0.066) & 0.398 (0.045) & \textbf{0.843} (0.072) \\
  & (standing, walking, elevator.up) & 0.468 (0.072) & 0.469 (0.024) & \underline{0.830} (0.038) \\
  & (running, elevator.down, throwing, bus) & \underline{0.613} (0.055) & \underline{0.379} (0.024) & 0.802 (0.081) \\
  & (stair.up, escalator.down, escalator.up, car) & 0.558 (0.041) & 0.419 (0.044) & \textbf{0.858} (0.043) \\

\hline
\multirow{7}{*}{\modelname} 
  & (sitting, bus) & \textbf{0.717} (0.089) & \textbf{0.313} (0.066) & \textbf{0.841} (0.052) \\
  & (stair.down, dragging, throwing) & \underline{0.749} (0.087) & \underline{0.303} (0.078) & \textbf{0.837} (0.043) \\
  & (sitting, running, bus) & \underline{0.608} (0.066) & \underline{0.397} (0.041) & \textbf{0.848} (0.056) \\
  & (cycling, kicking, tram) & \textbf{0.766} (0.077) & \textbf{0.303} (0.060) & \underline{0.837} (0.030) \\
  & (standing, walking, elevator.up) & \textbf{0.829} (0.081) & \textbf{0.232} (0.069) & \textbf{0.854} (0.039) \\
  & (running, elevator.down, throwing, bus) & \textbf{0.662} (0.081) & \textbf{0.358} (0.052) & \textbf{0.844} (0.039) \\
  & (stair.up, escalator.down, escalator.up, car) & \underline{0.673} (0.085) & \underline{0.359} (0.052) & \textbf{0.870} (0.038) \\
  
\hline
\end{tabular}
}
\caption{OOD results when multiple randomly chosen classes are included in the OOD set. Best results are in bold. Underlined scores indicate models whose performance is not statistically significantly different (p $\geq$ 0.05) from the best-performing model, according to a paired t-test over the folds.}
\label{tab:big_set_ood_results}
\end{table}

\end{document}